\documentclass{article}

\usepackage{microtype}
\usepackage{graphicx}
\usepackage{subcaption}

\usepackage{longtable}
\usepackage{tabularx}
\usepackage{booktabs} 
\usepackage{xcolor}
\definecolor{darkgreen}{rgb}{0.13, 0.55, 0.13}

\usepackage{hyperref}

\usepackage[preprint]{icml2026}

\usepackage{amsmath}
\usepackage{amssymb}
\usepackage{mathtools}
\usepackage{amsthm}

\usepackage[capitalize,noabbrev]{cleveref}

\theoremstyle{plain}

\theoremstyle{definition}

\theoremstyle{remark}

\usepackage[textsize=tiny]{todonotes}

\icmltitlerunning{Group-Adaptive Threshold Optimization for Robust AI-Generated Text Detection}

\newcommand{\name}{FairOPT}

\begin{document}

\twocolumn[
  \icmltitle{Group-Adaptive Threshold Optimization for \\
  Robust AI-Generated Text Detection}



  \icmlsetsymbol{equal}{*}

    \begin{icmlauthorlist}
        \icmlauthor{Minseok Jung}{mit}
        \icmlauthor{Cynthia Fuertes-Panizo}{paristech}
        \icmlauthor{Liam Dugan}{upenn}
        \icmlauthor{Yi R. Fung}{mit}
        \icmlauthor{Pin-Yu Chen}{ibm}
        \icmlauthor{Paul Pu Liang}{mit}
    \end{icmlauthorlist}

    \icmlaffiliation{mit}{Massachusetts Institute of Technology, USA}
    \icmlaffiliation{upenn}{University of Pennsylvania, USA}
    \icmlaffiliation{paristech}{Institut Polytechnique de Paris}
    \icmlaffiliation{ibm}{IBM Research, USA}
    
    \icmlcorrespondingauthor{Minseok Jung}{msjung@mit.edu}

    \icmlkeywords{detection, robustness, fairness, threshold optimization}

  \vskip 0.3in
]



\printAffiliationsAndNotice{}  

\begin{abstract}
    The advancement of large language models (LLMs) has made it difficult to differentiate human-written text from AI-generated text. AI text detectors have been developed in response, which typically utilize a fixed global threshold (e.g., $\theta = 0.5$) to classify AI-generated text. However, one universal threshold could fail to account for distributional variations by subgroups. For example, when using a fixed threshold, detectors make more false positive errors on shorter human-written text. These discrepancies can lead to misclassifications that disproportionately affect certain groups. We address this critical limitation by introducing \emph{\name}, an algorithm for group-specific threshold optimization for probabilistic AI-text detectors. We partitioned data into subgroups based on attributes (e.g., text length and writing style) and implemented \textit{\name} to learn decision thresholds for each group to reduce discrepancy. Across nine detectors and three heterogeneous datasets, our method reduces disparity by 27.4\% over five fairness metrics, while incurring less than 0.1\% sacrifice in accuracy, compared to static and ROC-based thresholding baselines. Our framework paves the way for more robust classification in AI-generated content detection via post-processing. We release our data, code, and project information at \href{https://anonymous.4open.science/r/threshold_optimization-BC9C/README.md}{URL}.  
\end{abstract}


\section{Introduction}

\begin{figure}[t]
    \centering    
    \includegraphics[width=\columnwidth]{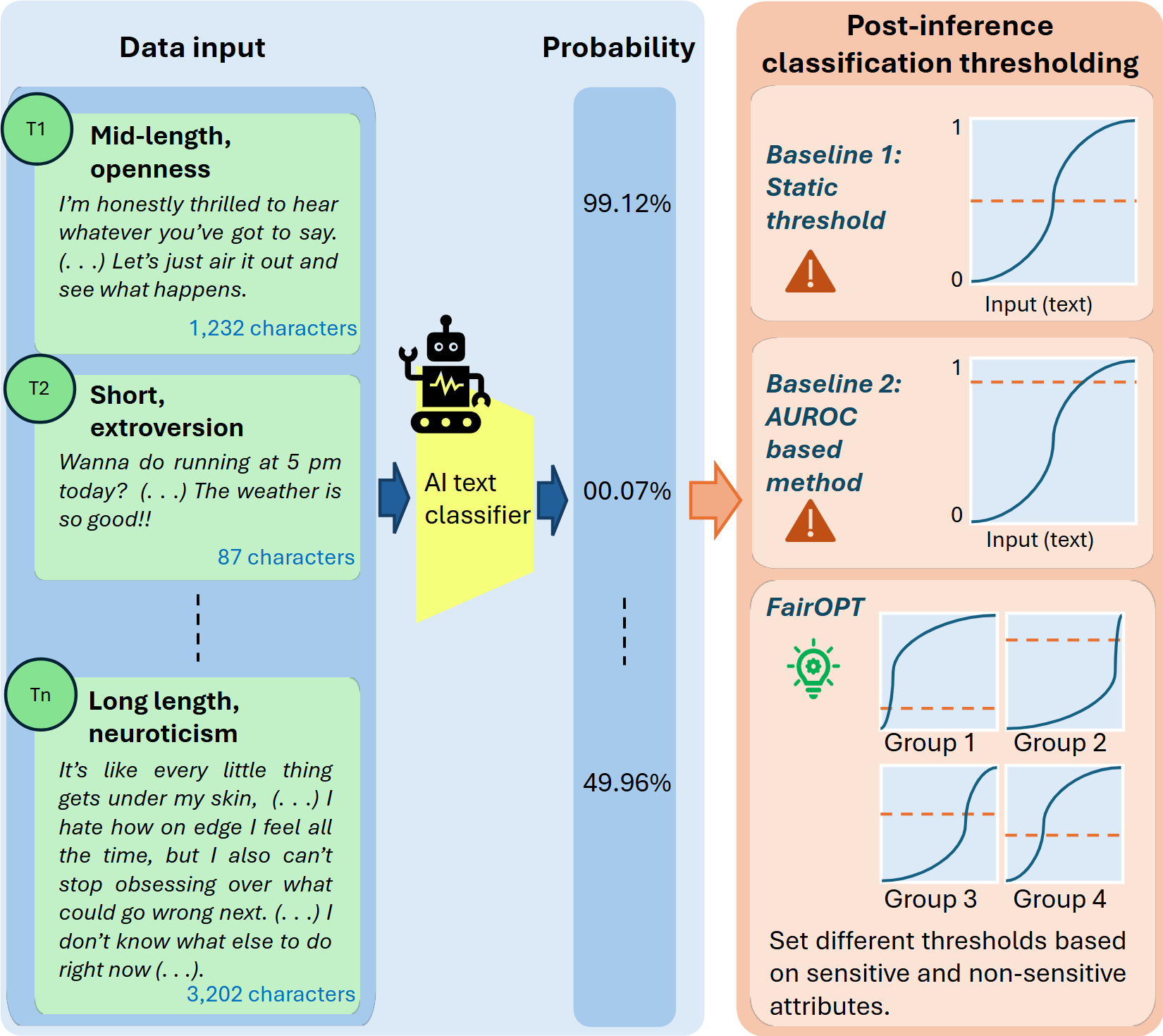}
    \caption{Using a single probability threshold (often $\theta=0.5$) for AI text classification ignores subgroup-specific score distributions (e.g., by length or style), increasing error rates for certain demographics. We propose \emph{\name}, which learns subgroup-specific decision thresholds that reduce these disparities (Figure~\ref{fig:result_discrepancy}) while incurring negligible drop in accuracy (Figure~\ref{fig:result_accuracy}).}
    \label{fig:fig1}
\end{figure}

\begin{figure*}
    \centering
    \includegraphics[width=0.97\textwidth]{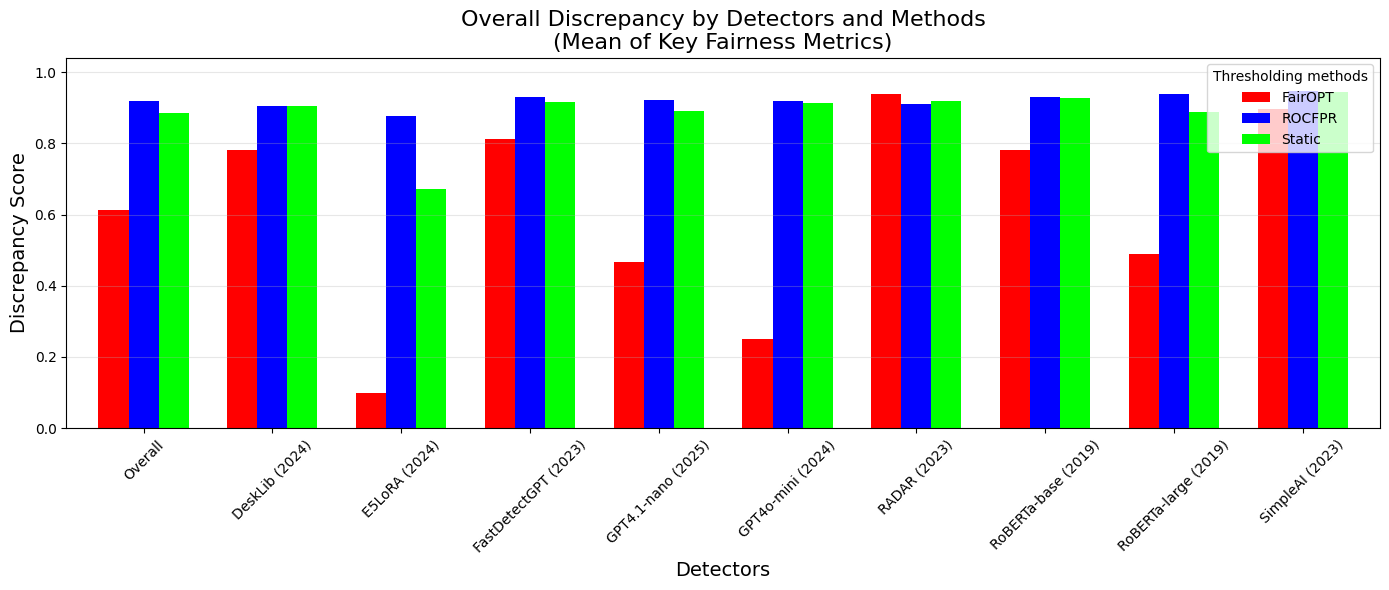}
    \caption{\emph{\name} decreases  overall discrepancy of AI detectors compared to other thresholding methods. FairOPT reduces average classification discrepancy of five fairness metrics to $\mu=0.613$, outperforming ROCFPR ($\mu=0.920$) and Static ($\mu=0.887$). Each bar shows average of five key fairness metrics that has been introduced in \ref{fairness_metrics}.}
    \label{fig:result_discrepancy}
\end{figure*}

The widespread adoption of large language models (LLMs) has led to a massive increase in AI-generated text~\cite{wu2025survey, guo2024biscope}, raising concerns about misinformation~\cite{chen2024combating}, publication standards~\cite{wu2023survey}, cybersecurity~\cite{yao2024survey}, and academic integrity~\cite{perkins2023academic}. To mitigate these risks, a variety of AI-text detection methods have been proposed~\cite{wu2025survey, bhattacharjee2024fighting}. These methods typically assign a probability that a given text is AI-generated by comparing its statistical patterns (e.g., token probabilities) to those of AI-generated text, and then apply a single universal decision threshold (e.g., $\theta=0.5$)~\cite{freeman2008comparison}  to classify text as AI-generated or human-written.

However, a universal threshold ignores the heterogeneity of the probability distributions across different textual characteristics, such as length and writing style. For example, long, highly neurotic prose may be more likely to be flagged as AI-generated than text with high agreeableness, leading to disproportionate misclassification rates for certain groups. Similar disparities have also been documented in classification involving race~\cite{alghamdi2022beyond, bendekgey2021scalable} and gender~\cite{weber2020black, jang2022group, bendekgey2021scalable}.

To address these limitations, we introduce \textit{\name}, a group-adaptive thresholding framework that learns distinct decision thresholds for predefined subgroups instead of relying on a single global threshold (see Figure~\ref{fig:fig1}). Our method jointly optimizes performance (e.g., accuracy, F1) and fairness (e.g., demographic parity, equality of odds) through the learning of group-specific thresholds. Experiments on nine AI text detectors across three benchmark datasets show that \textit{\name} substantially reduces disparities while incurring only a marginal loss in overall performance.

Our main contributions are threefold: (a) we identify substantial disparities in AI-text detectors across text length and stylistic characteristics; (b) we propose \textit{\name}, a group-adaptive thresholding framework that calibrates decision boundaries to mitigate these disparities; and (c) we demonstrate that \textit{\name} achieves a 27.4\% reduction in subgroup discrepancy across five fairness metrics, while incurring less than 0.1\% of sacrifice in overall accuracy.

\begin{figure*}
    \centering
    \includegraphics[width=0.92\textwidth]{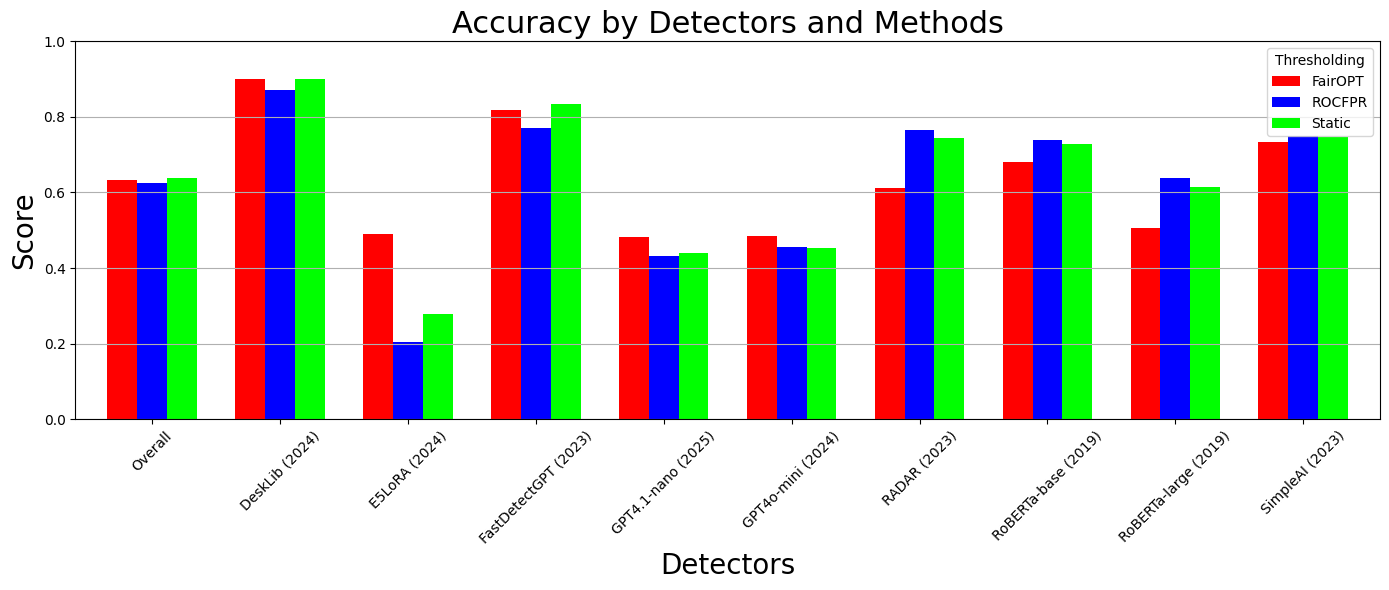}
    \caption{\textit{\name} shows a negligible sacrifice of ACC by effectively handling the performance-fairness tradeoff using a relaxed fairness rule that is shown in Figure \ref{fig:tradeoff}. Our method incurs minimal loss of ACC, achieving $\mu=0.6344$ compared to Static ($\mu=0.6376$) and ROCFPR ($\mu=0.6249$) methods. The accuracy metric is defined in Appendix~\ref{app:metrics}}
    \label{fig:result_accuracy}
\end{figure*}


\section{Related Work}

\subsection{AI text detectors}

AI text detectors have been developed to mitigate risks in misinformation, security, and integrity. RADAR leverages adversarial learning techniques \cite{hu2023radar}, GLTR uses metrics such as entropy and token-level probability ranks \cite{gehrmann2019gltr}, DetectGPT exploits local curvature of log probabilities under the reference LLM \cite{mitchell2023detectgpt}, and RoBERTa-based detectors fine-tune pretrained language models \cite{solaiman2019release}. In this work, we focus on \emph{probabilistic classifiers} that output a probability score and apply a decision threshold to obtain a binary label.

\subsection{Static thresholding for classification}\label{method1:static} 

The canonical choice in probabilistic classification is the threshold of $\theta=0.5$ to map probabilities to labels \cite{freeman2008comparison}. This default rule is widely used in many domains such as photogrammetry \cite{shao2016characterizing}, ecology \cite{manel1999comparing, hanberry2013prevalence}, and machine learning \cite{lu2024mlnet}. Nevertheless, relying solely on $\theta=0.5$, purely by convention, could be unreliable when data distributions vary \cite{freeman2008comparison}.

\subsection{Universal thresholding with optimization}\label{method2:auroc}

Conventional threshold optimization methods learn a single threshold over the dataset by optimizing performance metrics such as true positive rate (TPR) and false positive rate (FPR) \cite{krishna2024paraphrasing, lipton2014optimal}. A prominent approach selects the threshold using the receiver operating characteristic curve (AUROC) \cite{bradley1997use}. Similarly, ROCFPR maximizes TPR subject to low FPR (e.g., $\leq 0.1$) along the same ROC curve.  These methods are used in econometrics \cite{stavnkova2023threshold}, statistics \cite{esposito2021ghost}, and AI \cite{openai2023classifier}. While AUROC-based optimization yields a globally tuned threshold for a given model, it may overfit or fail to shift to another environment.

\subsection{Adaptive thresholding across groups}

Several works advocate group-adaptive thresholds to reduce disparities \cite{jang2022group, bakker2021beyond}. Instance-dependent thresholds for predicted probabilities have been proposed to reduce error discrepancies \cite{menon2018cost}. Group-specific thresholds have been shown to support fairness-aware classification \cite{corbett2017algorithmic}, and race-specific thresholds can reduce performance gaps \cite{canetti2019soft}. Other methods adapt thresholds to ensure fair classification \cite{jang2022group} or to stabilize performance over time \cite{bakker2021beyond}. Building on these foundations, our framework \emph{\name} learns group-adaptive thresholds that jointly optimize performance and fairness.


\section{Conceptual Framework and Methodology} \label{sec:methods}

This section outlines our method \textit{\name} for adaptive thresholding, which iteratively adjusts classification thresholds for different subgroups to reduce discrepancy.


\subsection{Objective}

Consider a set of subgroups $G_1,\ldots,G_i$ defined by attributes (e.g., length, style). Acknowledging that texts with varying characteristics exhibit different probability distributions, our objective is to find optimal thresholds $\{\theta(G_1),\ldots,\theta(G_i)\}$ for binary classification.


\subsubsection{Preliminaries of AI Text Detection}

The detection of AI-generated text is commonly formulated as a probabilistic binary classification task, where the input space $\mathcal{X}$ is a sequence of tokens and the label space $\mathcal{Y}=\{0,1\}$ denotes human-written ($0$) versus AI-generated ($1$). Existing work primarily uses a pretrained neural network as a probabilistic classifier $M_\theta:\mathcal{X}\to[0,1]$ that assigns to each input a probability $p$ of being AI-generated \cite{solaiman2019release, hu2023radar}. This probability is then converted into the predicted label $\hat{Y}\in\{0,1\}$ by applying a threshold, typically $0.5$:

\begin{align}
    \hat{Y} = \begin{cases}
        1, & \text{if } M_\theta(x) \geq \text{threshold}, \\
        0, & \text{otherwise.}
        \end{cases}
\end{align}


\subsubsection{Performance metrics} 

After inferring $\hat{Y}$, we compare $Y$ and $\hat{Y}$. Performance and fairness metrics details are in Appendix~\ref{app:performance_fairness_metrics}.


\subsubsection{Minimax problem}

We quantify subgroup disparity across five complementary metrics: Demographic Parity (DP)~\cite{kim2020fact}, Balanced Error Rate (BER)~\cite{ferrer2022analysis}, Equalized Odds, FPR, and Predictive Parity. For metric $k$ and subgroup $g$, let $m_{k,g}$ denote the metric value. We define the inter-group discrepancy $\Delta_k$ and formulate our minimax objective to ensure fairness holds for the \emph{worst-case subgroup} rather than merely on average, thereby controlling tail risk:

\begin{equation}
    \Delta_k = \max_g m_{k,g} - \min_g m_{k,g}
\end{equation}

\begin{equation}
    \text{Objective:}\quad
    \min_{\{\theta_g\}} \max_k \Delta_k(\{\theta_g\})
\end{equation}


\subsubsection{Fairness metrics}\label{fairness_metrics}

We employ five complementary fairness metrics, each capturing different aspects of group equity. 

For each metric, perfect fairness is achieved when the disparity $\Delta_{\text{metric}}$ is $0$; relaxed formulations allow $\Delta_{\text{metric}} \geq 0$ for practical feasibility.


\paragraph{Demographic Parity (DP).} DP seeks to measure \emph{disparate impact}, where one group receives disproportionate results \cite{feldman2015certifying}. Requires statistically independent across attributes: $\hat{Y} \perp\!\!\!\perp S$. This condition mandates that the probability of assigning a positive outcome is equal across all groups. Perfect 
fairness is achieved when the discrepancy $\Delta_{\text{DP}}$ 
reaches $0$, where:

\begin{align}
    \Delta_{\text{DP}} &= \left| P(\hat{Y} = 1 \mid S = a) - P(\hat{Y} = 1 \mid S = b) \right|.
\end{align}


\paragraph{Balanced Error Rate (BER).} BER equally weights 
Type I and Type II errors, capturing both false positives and false negatives~\cite{ferrer2022analysis}. We compute BER for each subgroup $G_i$ and measure discrepancy as follows, where lower discrepancy indicates more equitable error rates across groups:

\begin{equation}
    \mathrm{BER}_i = \tfrac{1}{2}\Bigl(\underbrace{\tfrac{\mathrm{FP}_i}{\mathrm{FP}_i+\mathrm{TN}_i}}_{\mathrm{FPR}_i}\; +\; \underbrace{\tfrac{\mathrm{FN}_i}{\mathrm{FN}_i+\mathrm{TP}_i}}_{\mathrm{FNR}_i}\Bigr)  
\end{equation}

\begin{equation}
    \Delta_{\text{BER}} = \max_i \mathrm{BER}_i - \min_i \mathrm{BER}_i    
\end{equation}


\paragraph{Equalized Odds (EO).} EO requires that a classifier's predictions are conditionally independent of attributes $S$ given the true label $Y$~\cite{hardt2016equality}. This ensures equal TPR and FPR across groups, treating individuals similarly in both favorable outcomes (true positives) and unfavorable outcomes (false positives). The discrepancy is defined as the maximum gap in either metric across subgroups, where perfect fairness is 
achieved when $\Delta_{\text{EO}} = 0$:

\begin{align}
    \Delta_{\text{EO}}
    &= \max\Bigl(
    \max_{g} \text{TPR}_g - \min_{g} \text{TPR}_g,\;
    \max_{g} \text{FPR}_g - \min_{g} \text{FPR}_g
    \Bigr).
\end{align}


\paragraph{False Positive Rate (FPR).} FPR focuses on subgroup differences in false alarm rates, when the classifier predicts $\hat{Y}=1$ despite true label $Y=0$. This metric is critical in high-stakes applications where incorrect positive predictions disproportionately harm certain groups. The discrepancy is defined as:

\begin{equation}
    \Delta_{\text{FPR}} = \max_{g} \text{FPR}_g - \min_{g} \text{FPR}_g
\end{equation}


\paragraph{Predictive Parity (PP).} PP measures whether a  positive prediction carries the same meaning across groups. I.e. whether $P(Y=1 \mid \hat{Y}=1, S=g)$ is group-independent. This ensures precision (true positives among predicted positives) is consistent across subgroups:

\begin{align}
\Delta_{\text{PP}} &= \max_{g} \text{PP}_g - \min_{g} \text{PP}_g
\quad , \quad
\text{PP}_g = \frac{\text{TP}_g}{\text{TP}_g + \text{FP}_g}
\end{align}


\subsection{Relaxed Fairness}
\label{sec:fairness_criterion}

While achieving perfect fairness across groups (e.g., $\Delta_{\text{k}} = 0$) is theoretically ideal, it is often impractical due to computational constraints and the difficulty of training models under strict fairness constraints~\cite{dwork2012fairness}. The 80\% relaxed fairness rule~\cite{feldman2015certifying} offers a pragmatic alternative (e.g., allowing $\Delta_{\text{DP}} = 0.2$), providing a more feasible fairness criterion and reducing overfitting risk.

We formalize a relaxed fairness criterion requiring that the ratio of the minimum to maximum value of each fairness metric across all subgroups does not fall below a flexibility parameter $\tau = 0.8$:

\begin{align}
    \frac{\min\limits_{i \in \mathcal{S}} \, c_i^{(k)}(\theta)}{\max\limits_{i \in \mathcal{S}} \, c_i^{(k)}(\theta)} \geq \tau, \quad \forall k \in \mathcal{K},
\end{align}

where:

\begin{itemize}
    \item $\mathcal{S}$ is the set of subgroups defined by text attributes (e.g., length, writing style).
    
    \item $\mathcal{K}$ is the set of fairness metrics (e.g., DP, EO, BER).
    
    \item $c_i^{(k)}(\theta)$ is the value of metric $k \in \mathcal{K}$ for subgroup $i \in \mathcal{S}$ under classifier parameters $\theta$.
    
    \item $\tau \in [0,1]$ is the fairness parameter; $\tau = 1$ corresponds to the strictest notion of fairness (perfect fairness).
\end{itemize}

We restrict the set of fairness metrics in light of the \emph{impossibility theorem}~\cite{kleinberg2017inherent} and to maintain computational feasibility, since satisfying many fairness criteria simultaneously is impractical (Appendix \ref{sec:impossibility}).

\subsection{Group-Adaptive Threshold Optimization} \label{subsec:group_adaptive}


\begin{algorithm}[t]
\caption{\textbf{FairOPT: Group-Specific Thresholding}} 
    
    \begin{algorithmic}[1]
    \STATE \textbf{Input:} 
        \(D\) dataset, 
        \(S_n\) features,
        predicted probabilities \(p\),  
        performance targets (\(\alpha,\beta\)), 
        fairness gap limit \(\delta_{\text{fair}}\), 
        max iterations, step size \(\eta\).
    
    \STATE Split \(D\) into sub-datasets \(D_1,\ldots,D_j\) by \(S_n\).
    \FOR{\text{iteration} = 1 to \text{max iterations}}
        \FOR{each subgroup \(G_i\)}
            \STATE Convert \(p\) in \(G_i\) to binary predictions using \(\theta(G_i)\) and compute confusion matrix.
            \IF{(\(\mathrm{ACC}_i \ge \alpha\) and \(\mathrm{F1}_i \ge \beta\) for all \(i\)) \textbf{and} (fairness gap \(\le \delta_{\text{fair}}\))}
                \STATE \textbf{return} \(\{\theta(G_1),\ldots,\theta(G_j)\}\).
            \ELSE 
                \STATE Adjust \(\theta(G_i)\) 
            \ENDIF
        \ENDFOR
    \ENDFOR
    \STATE \textbf{Output:} \(\{\theta(G_1),\dots,\theta(G_j)\}\).
    \end{algorithmic}
    
\label{algorithm_fairopt}
\end{algorithm}



\paragraph{Process.} Given this setup, we introduce \textit{\name}, described in Algorithm~\ref{algorithm_fairopt}. \textit{\name} uses a gradient-based subgroup threshold update to jointly optimize group-level fairness and classification performance. Formally, we partition the dataset \(D\) into subgroups \(\{G_1,\dots,G_i\}\) based on features \(\{S_1,\dots,S_n\}\). Each subgroup \(G_i\) is assigned an initial threshold \(\theta_{\mathrm{init}}\) (e.g., \(0.5\) for all groups). At each iteration, we compute predictions for each sample \(x_n \in G_i\) by binarizing its predicted probability \(p_n\) using the current threshold \(\theta(G_i)\). From the predicted labels \(\hat{y}_n\) and true labels \(y_n\), we construct a contingency matrix. For each subgroup \(G_i\), we define a loss that penalizes low accuracy and insufficient F1 score:

\begin{align}
   L_i(\theta) 
   \;=\; 
   -\,\mathrm{ACC}_i(\theta)
   \;+\;
   \kappa \,\bigl[\beta - \mathrm{F1}_i(\theta)\bigr]_{+},
\end{align}

where \(\beta\) is a minimum F1 target, \(\kappa>0\) is a penalty weight, and \([\cdot]_{+}\) denotes the positive part, which clips negative values to zero. These minimum standards ensure stable convergence and prevent degenerate solutions across subgroups.

To optimize each \(\theta(G_i)\), we employ a \emph{finite-difference approximation} of the gradient:

\begin{align}
   \nabla L_i(\theta(G_i)) 
   \;\approx\;
   \frac{L_i(\theta(G_i)+\delta) - L_i(\theta(G_i)-\delta)}{2\,\delta}.
\end{align}

This estimates the gradient of the subgroup loss by perturbing the threshold by \(\pm\delta\) and taking the symmetric difference quotient~\cite{liu2020primer}. We then perform a gradient-descent update for each subgroup, followed by clipping to \([a,b]\subseteq[0,1]\) to ensure a valid probability.

\begin{align}
   \theta(G_i) 
   \;\leftarrow\;
   \theta(G_i) - \eta\, \nabla L_i(\theta(G_i)),
\end{align}

Finally, the model evaluates the fairness metrics \(\{M_1, \ldots, M_k\}\) across all subgroups via the inter-group disparity, and checks whether \(\Delta_k \le \delta_{\mathrm{fair}}\) for each metric \(M_k\). If any disparity exceeds \(\delta_{\mathrm{fair}}\) (e.g., \(0.2\)), the model is considered unfair. This fairness test is kept outside the main loss because it is a binary pass–fail criterion.

\begin{align}
    \Delta_{k} \;=\; \max_i\,M_k(G_i) \;-\;\min_i\,M_k(G_i),
\end{align}

To determine termination, we monitor the largest threshold change among all groups. If \(\Delta_\theta\) falls below a tolerance \(\epsilon_{\mathrm{tol}}\) and all subgroups satisfy the performance criteria (\(\mathrm{ACC}_i \ge \alpha\), \(\mathrm{F1}_i \ge \beta\)) and fairness constraints (\(\Delta_k \le \delta_{\mathrm{fair}}\)), the algorithm terminates.

\begin{align}
   \Delta_\theta \;=\; \max_{G_i}\,\bigl|\theta_{\mathrm{new}}(G_i) - \theta_{\mathrm{old}}(G_i)\bigr|.
\end{align}

\paragraph{Outcome} 
By iterating this process, the Algorithm \ref{algorithm_fairopt} converges to a set of subgroup thresholds \(\{\theta(G_1),\dots,\theta(G_i)\}\) that reduces the disparity while satisfying the demanded performance criteria. For full implementation details and hyperparameter specifications, refer to Appendix~\ref{appendix:algorithm}.


\section{Experiments}

Our experiments evaluate \textit{\name} against static and AUROC-based threshold optimization methods. We apply the feature engineering described in Section~\ref{attributes} and the AI detectors in Section~\ref{ai_classifiers} to extract features and obtain classifier predictions.


\subsection{Dataset}

The dataset is constructed from RAID~\cite{dugan-etal-2024-raid}, MAGE~\cite{li-etal-2024-mage}, and SemEval24~\cite{wang2024semeval}, with an even balance between human and AI text to avoid class imbalance. The AI-generated texts include GPT and Text-Davinci variants, and the corpus spans diverse domains such as social media (e.g., Reddit, ELI5), encyclopedic and academic sources (e.g., Wikipedia, arXiv), instructional and creative content (e.g., WikiHow, poetry, recipes), and reviews and summaries (e.g., Yelp, XSum), covering both formal and informal styles. The dataset comprises 19{,}117 training instances and 4{,}779 test samples.


\subsection{AI-Generated Text Classifiers}
\label{ai_classifiers}

To classify AI-generated text, we employed both open-source and closed-source detectors. The open-source models are DeskLib~\cite{desklib-ai-text-detector-2024}, E5LoRA~\cite{zhou2024e5smalllora}, FastDetectGPT~\cite{bao2023fast}, RADAR~\cite{hu2023radar}, RoBERTa-base and -large~\cite{solaiman2019release}, and SimpleAI's detector~\cite{guo-etal-2023-hc3}. Whereas, the closed-source models are GPT4.1-nano and GPT4o-mini~\cite{openai-gpt4.1-nano-2025, openai-2024-gpt4o-mini}, which have been used for LLM-based text detection~\cite{bhattacharjee2024fighting}. The open-source detectors are used for both training and testing, while the closed-source models are applied only at test time.


\subsection{Subgroup Attributes}
\label{attributes}

We extract four attributes for subgrouping (Table~\ref{tab:featuretable}). These comprise one non-sensitive structural attribute and three sensitive writing attributes.

\begin{table}[h!]
  \small
  \centering
  \begin{tabularx}{\columnwidth}{@{} l X X @{}}
    \toprule
    \textbf{Attribute} & \textbf{Characteristics} & \textbf{Labels} \\
    \midrule[0.8px]
    
    Text length
      & Number of characters in text. \newline ($k = 3$)
      & Short ($\leq1K$), Medium ($\leq 2.5K$), Long ($> 2.5K$) \\      
    \midrule
    
    Writing style
      & Inferred Big Five personality traits. \newline ($k = 5$)
      & Openness,\newline Conscientiousness,\newline Extraversion,\newline Agreeableness,\newline Neuroticism \\ 
    \midrule
    
    Formality
      & Formal expression. \newline ($k = 2$)
      & Formal,\newline Informal \\ 
    \midrule
    
    Sentiment
      & Binary sentiment. \newline ($k = 2$)
      & Positive,\newline Negative \\
      
    \bottomrule
  \end{tabularx}
  \caption{Subgroup attributes from the feature engineering.}
  \label{tab:featuretable}
\end{table}

The \textbf{text length} attribute captures document length. Building on evaluations revealing performance disparities~\cite{openai2023classifier} and exploratory data analysis, we establish categorical distinctions based on character count. On the other hand, the three sensitive writing attributes are: \textbf{writing style}, derived from a BERT-based personality classifier~\cite{personality_prediction}; \textbf{formality}, computed using the Heylighen \& Dewaele formality score~\cite{Formality_Heylighen2002}; and \textbf{sentiment}, obtained via DistilBERT SST-2 classification~\cite{DistilBERTAD2019}.


\subsection{Discovery of Subgroup Bias in Text Detection}
\label{sec:bias_discovery}

\begin{figure}
  \centering
  \includegraphics[width=0.48\textwidth]{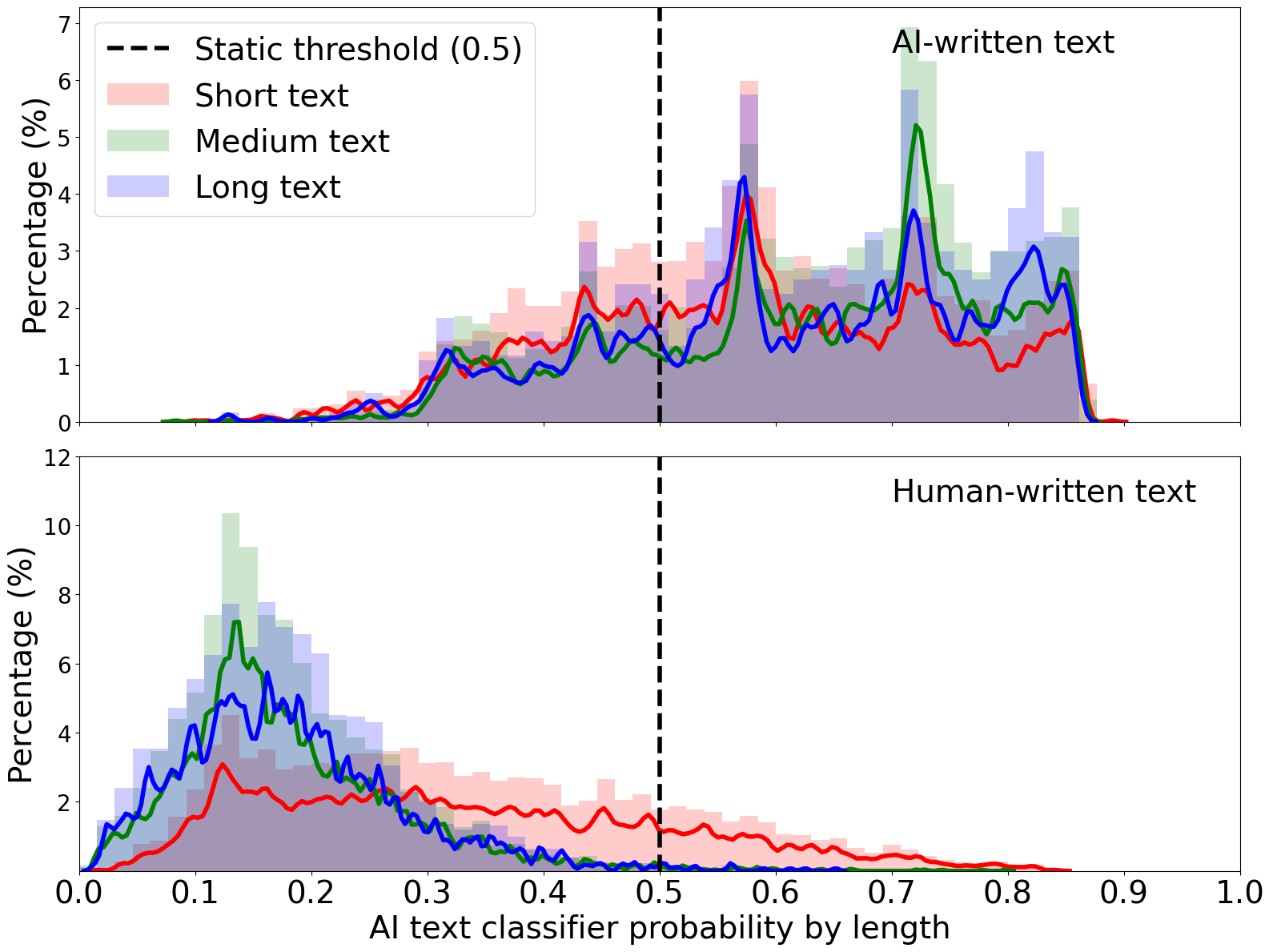}
  \caption{Human-written short texts exhibit higher FPR than other text lengths under static thresholding ($\theta=0.5$). Curves show mean AI-likelihood scores across detectors in the training dataset.}
  \label{fig:text_probability_distribution}
\end{figure}

\begin{figure*}
    \centering
    \includegraphics
    [width=0.97\textwidth]{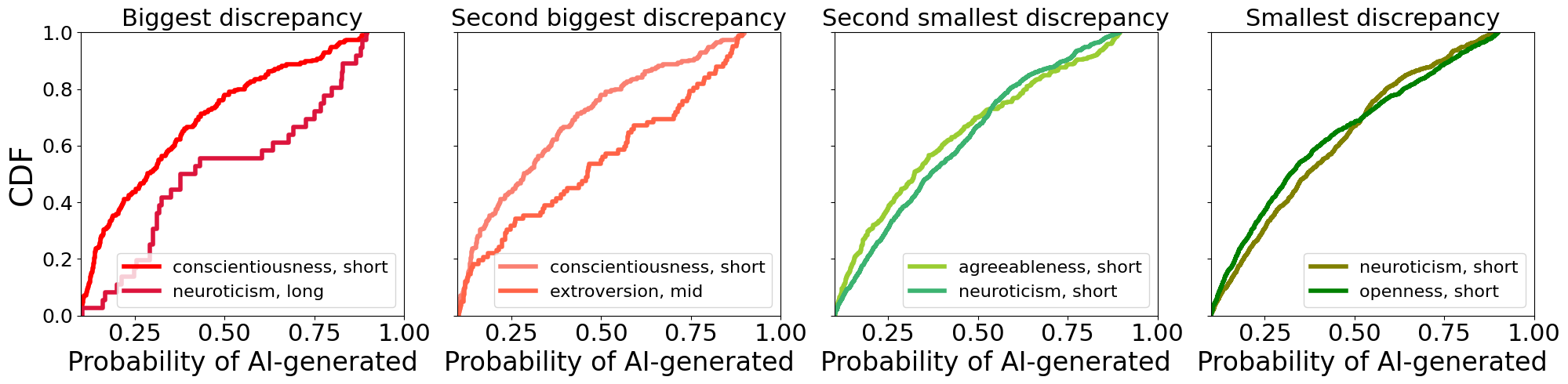}
    \caption{Kolmogorov–Smirnov test comparing score distributions across subgroup pairs (RoBERTa-large). Shown are the two largest and two smallest distributional differences. Largest discrepancy: conscientiousness-short vs. neuroticism-long ($KS = 0.3081$, $p < 0.01$). Second largest: conscientiousness-short vs. extroversion-medium ($KS = 0.2616$, $p < 0.01$). Second smallest: agreeableness-short vs. neuroticism-short ($KS = 0.0909$, $p < 0.05$). Smallest: neuroticism-short vs. openness-short ($KS = 0.0842$, $p < 0.01$).}
    \label{fig:discrepancy_two_features}
\end{figure*}

We assess detection performance across text-length subgroups using static thresholding ($\theta=0.5$). As shown in Figures~\ref{fig:text_probability_distribution}, short human-authored texts exhibit substantially elevated false positive rates compared to longer texts. Additionally, to examine the combined effect of length and writing style, we conduct a two-sample Kolmogorov-Smirnov test across subgroup pairs (Figure~\ref{fig:discrepancy_two_features}), confirming statistically significant distributional divergence ($KS_{\max}=0.3081$, $p<0.01$). These disparities systematically arise under static thresholding, motivating group-adaptive approaches. 


\subsection{FairOPT for Bias Mitigation} \label{sec:fairopt_method}

\begin{figure} [H]
  \centering
  \includegraphics[width=0.45\textwidth]{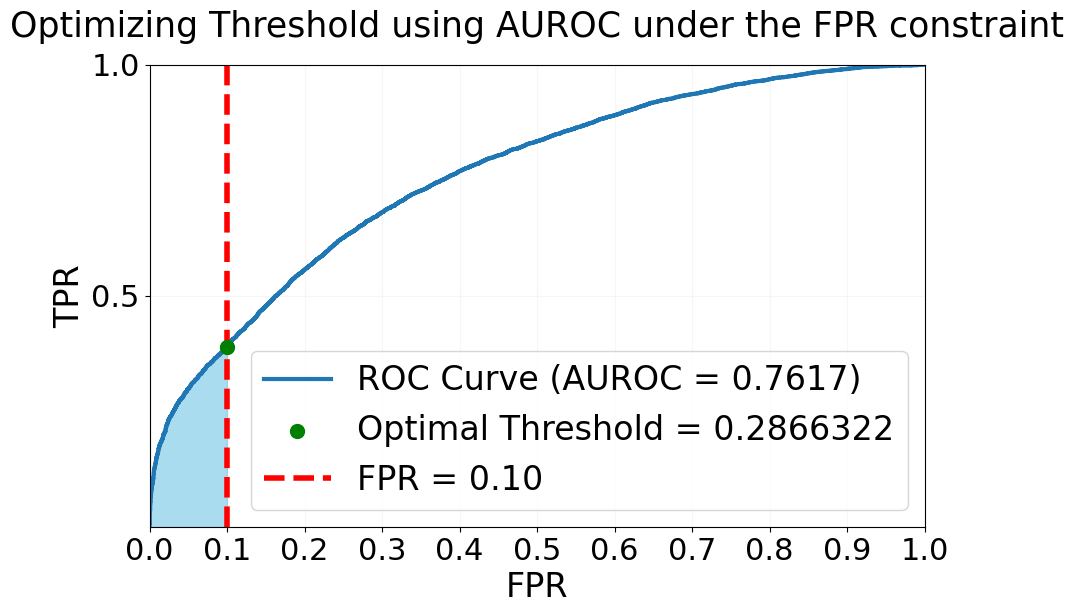}
    \caption{Threshold optimization based on the AUROC‐based method under an $FPR \le 10\%$ constraint. The ROC curve (blue line) illustrates the trade‐off between TPR and FPR, and the optimal threshold (green marker) maximizes TPR under the specified FPR constraint.}
  \label{fig:aurocfpr}
\end{figure}

We apply \textit{\name} to learn subgroup-specific thresholds $\{\theta(G_j)\}_{j=1}^{N}$, generating decision thresholds (N = 60) to each group (Appendix \ref{app:applied_thresholds} includes N thresholds that have been generated via FairOPT). Using predicted probabilities $p \in [0,1]$ from multiple AI text classifiers (Section~\ref{ai_classifiers}), FairOPT optimizes RoBERTa-large predictions~\cite{solaiman2019release} on the training set under relaxed fairness constraints ($\tau=0.8$; Section~\ref{sec:fairness_criterion})~\cite{wang2023aleatoric}. Test set evaluation reports discrepancy, accuracy, and F1 score (Appendix~\ref{app:metrics}). As baselines, we are using Static ($\theta=0.5$) and AUROC-optimized thresholds (see Figure~\ref{fig:aurocfpr}).

\subsection{Tradeoff Analysis between Performance and Fairness}

\begin{figure}[H]
  \centering
  \includegraphics[width=0.47\textwidth]{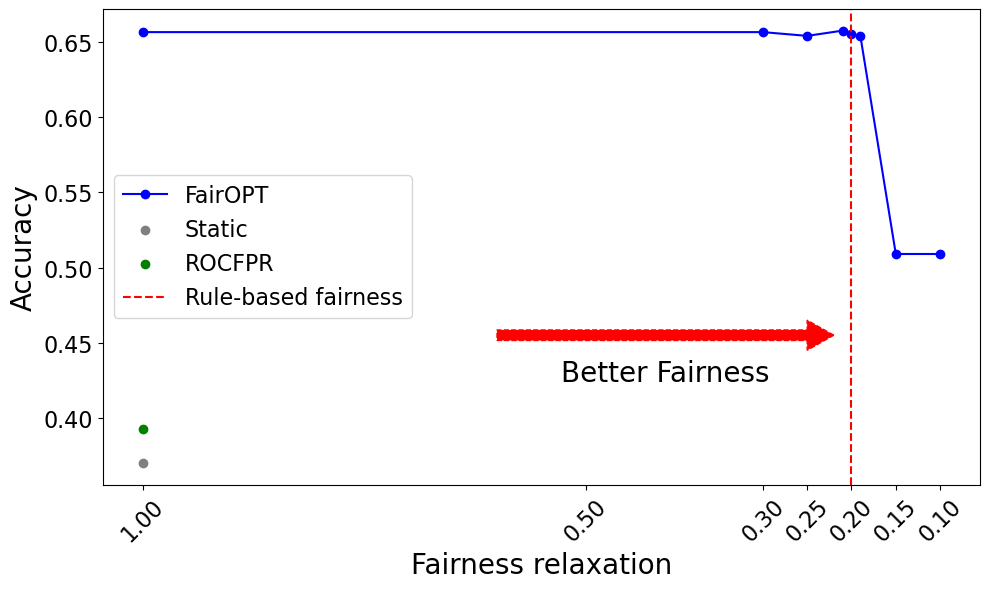}
  \caption{The \textit{\name} algorithm balances performance and fairness across relaxation levels: lower relaxation ($\tau \to 0$) prioritizes fairness, while higher relaxation ($\tau \to 1$) ignores fairness constraints between subgroups to optimize for accuracy. Static and AUROC-based methods occupy extremes; \textit{\name} maintains stable accuracy across fairness ranges. Details in Appendix~\ref{app:early_stopping}.
}
  \label{fig:tradeoff}
\end{figure}

Figure~\ref{fig:tradeoff} compares the performance-fairness trade-offs across \textit{\name}, static thresholding, and AUROC-based baselines. The relaxed fairness criterion (Section~\ref{sec:fairness_criterion}) enables \textit{\name} to reduce fairness discrepancy while preserving overall accuracy.

The AUROC-based method minimizes FPR in training at the cost of substantially reduced TPR (Figure~\ref{fig:aurocfpr}). In contrast, \textit{\name} achieves better fairness while maintaining competitive accuracy (Figure~\ref{fig:tradeoff}), demonstrating that subgroup-adaptive thresholding avoids the single-threshold trade-off.

\begin{table*}
    \centering
    \label{tab:fairopt-metrics}
    \begin{tabular}{lcccccc}
    \toprule
    \textbf{Metric} & \textbf{FairOPT} & \textbf{Static} & \textbf{$\Delta$ to Static} & \textbf{ROCFPR} & \textbf{$\Delta$ to ROCFPR} \\
    \midrule
    Accuracy                    & 0.6341 & 0.6376 & $-0.35\%$ & 0.6249 & $+1.47\%$ \\
    
    \midrule [0.3pt]
    Max BER Disparity         & \textbf{0.5852} & 0.7507 & {\color{darkgreen}$\downarrow$} $22.0\%$ & 0.7592 & {\color{darkgreen}$\downarrow$} $23.0\%$ \\
    
    Demographic Parity Disparity & \textbf{0.6772} & 0.9730 & {\color{darkgreen}$\downarrow$} $30.4\%$ & 0.9826 & {\color{darkgreen}$\downarrow$} $31.1\%$ \\
    
    Equalized Odds Disparity  & \textbf{0.6379} & 0.8501 & {\color{darkgreen}$\downarrow$} $24.9\%$ & 0.9198 & {\color{darkgreen}$\downarrow$} $30.6\%$ \\
    
    FPR Disparity             & \textbf{0.4313} & 0.9471 & {\color{darkgreen}$\downarrow$} $54.4\%$ & 0.9715 & {\color{darkgreen}$\downarrow$} $55.6\%$ \\
    
    Predictive Parity Disparity & \textbf{0.7333} & 0.9130 & {\color{darkgreen}$\downarrow$} $19.7\%$ & 0.9684 & {\color{darkgreen}$\downarrow$} $24.2\%$ \\
    \bottomrule
    \end{tabular}
    \caption{Fairness and performance metrics across methods. FairOPT substantially reduces fairness discrepancies (bold). $\Delta$ represents percentage-point reduction. Results aggregated across MAGE, RAID, and SemEval datasets using 60 subgroups (four attributes).}
    \label{tab:fairopt-metrics}
\end{table*}

\begin{table*}
    \centering
    \begin{tabular}{lcccccc}
    
    \toprule
    \textbf{Source} &
    \textbf{MaxBER Static} &
    \textbf{MaxBER FairOPT} &
    \textbf{$\Delta$ (\%)} &
    \textbf{Accuracy Static} &
    \textbf{Accuracy FairOPT} &
    \textbf{$\Delta$ (\%)} \\
    
    \midrule
    \textbf{MAGE}   & 0.7179 & \textbf{0.5283} & $-26.4$ & \textbf{0.8942} & 0.8872 & $-0.7$ \\
    
    \textbf{RAID}   & 0.6874 & \textbf{0.5404} & $-21.4$ & \textbf{0.9058} & 0.8978 & $-0.8$ \\
    
    \textbf{SemEval} & 0.6587 & \textbf{0.5221} & $-20.8$ & 0.8996 & \textbf{0.9036} & $+0.4$ \\
    
    \bottomrule
    \end{tabular}
    \caption{Per-dataset results: four-feature setting (60 subgroups).}
    \label{tab:datasets_results}
\end{table*}


\subsection{Results}
\label{sec:results}

\textbf{Decrease of the discrepancy.} 

Table~\ref{tab:fairopt-metrics} shows that group-adaptive thresholding with \textit{\name} consistently reduces subgroup discrepancy across detectors. Overall, our method decreased the average discrepancy around $27.4\%$, substantially reducing subgroup gaps compared to static and ROCFPR baselines. On average, \textit{\name} achieves a BER discrepancy of $0.5852$, representing a $22\%$ reduction relative to the static threshold baseline ($0.7507$) and a $23\%$ reduction compared to ROCFPR ($0.7592$). Similar improvements are observed across other metrics: demographic parity decreases from $0.9730$ to $0.6772$ ($30\%$), equalized odds from $0.8501$ to $0.6379$ ($25\%$), false-positive rate discrepancy from $0.9471$ to $0.4313$ ($51\%$), and predictive parity from $0.9130$ to $0.7333$ ($18\%$). Overall, \textit{\name} lowers the discrepancy. Full results are provided in Appendix~\ref{app:detailed_results}.


\textbf{Consistency across datasets.} 

FairOPT consistently reduces subgroup discrepancy across all evaluated fairness metrics on each dataset when compared to both Static and ROCFPR baselines (Table~\ref{tab:datasets_results}). Across MAGE, RAID, and SemEval, FairOPT lowers discrepancy of BER by 22\%, with corresponding reductions observed for DP by 30\%, EO by 25\%, FPR by 55\%, and PP by 20\%. Importantly, these fairness improvements are achieved with negligible impact on predictive performance: overall accuracy remains stable for FairOPT (0.634), closely matching Static (0.638) and exceeding ROCFPR (0.625).


\textbf{Detector-specific analysis.} 

Figure~\ref{fig:result_discrepancy} reveals strong improvements across all architectures. Overall, FairOPT achieves mean discrepancy $0.613$ versus $0.887$ (Static) and $0.920$ (ROCFPR). E5LoRA shows the largest gain: $0.100$ from $0.672$ (Static, $85\%$ reduction) or $0.878$ (ROCFPR, $89\%$ reduction). BERT-based classifiers also benefit substantially: RoBERTa-large drops to $0.490$ from $0.890$ (Static) and $0.940$ (ROCFPR), while RoBERTa-base falls to $0.782$ from $0.927$ (Static) and $0.932$ (ROCFPR). GPT-based detectors improve meaningfully (GPT4o-mini: $0.253$; GPT4.1-nano: $0.466$, compared to $0.9$ under both Static and ROCFPR), while smaller models show modest gains ($5\%$--$12\%$). RADAR is the sole exception, with negligible $2\%$ increase where ROCFPR underperforms Static. Accuracy remains stable across these improvements: FairOPT achieves $0.6341$, virtually identical to Static ($0.6376$, $-0.35\%$) and superior to ROCFPR ($0.6249$, $+1.47\%$), confirming that fairness gains do not sacrifice detection performance.


\subsection{Threshold Stability}
The stability of FairOPT were systematically evaluated across three highly diverse and representative benchmark datasets: MAGE, RAID, and SemEval. These datasets were intentionally selected to capture a broad spectrum of domains, genres, and writing styles, including academic and encyclopedic texts, social media posts, creative writing, and reviews. This diversity ensures that our subgroup definitions—based on features such as length, writing style, formality, and sentiment—are well supported and meaningful within each dataset. This provides a rigorous testbed for threshold optimization.


\section{Conclusions}
\label{sec:conclusion}


This paper proposes \textit{\name}, a post-processing algorithm for probabilistic classifiers that reduces discrepancy while preserving overall accuracy. We observe statistically significant differences in AI detector outputs across text characteristics (e.g., length and writing style), which cause disproportionately higher error rates for certain subgroups. Unlike universal thresholding methods, which assume evenly distributed probabilities regardless of stylistic features, \textit{\name} captures these distributional variations, achieving substantially lower classification discrepancy.


The novelty of this work lies in a group-adaptive, post-hoc threshold optimization framework that operates without retraining detectors, leveraging fairness and performance metrics. To our knowledge, no prior work has systematically examined threshold optimization for AI text detectors, despite existing methods for general probabilistic classifiers. Thus, \textit{\name} provides a practical alternative to single fixed thresholds (e.g., 0.5).


Although our experiments focused on English-language AI-generated text detectors, \textit{\name} is a modality-agnostic, subgroup-specific thresholding framework that readily extends to new deployment contexts and populations: for previously unseen subgroups, thresholds can be recalibrated from available data or warm-started from similar existing groups (e.g., AI-written text in schools, AI-generated images on a platform, or deepfakes on YouTube). Beyond text, \textit{\name} applies to image, video, and audio deepfake detection—any setting with probabilistic outputs that benefits from post-processing to ensure equitable performance—and more broadly to probabilistic classification tasks with subgroup disparities (e.g., recidivism risk prediction, hiring decisions), where learning tailored decision thresholds can mitigate bias with minimal impact on overall accuracy. We recognize that both theoretical analysis and user-centered case studies will be important for validating these extensions across diverse, real-world deployments.

\section*{Impact Statement}

\textit{\name} mitigates fairness disparities in AI-text detection through group-adaptive thresholding, enabling equitable classification across diverse text characteristics and user populations. This is critical for high-stakes applications including misinformation detection, academic integrity verification, and content moderation, where systematic errors disproportionately harm specific groups. By significantly reducing disparity across subgroups with minimal accuracy loss, our work promotes responsible deployment of probabilistic classifiers in real-world detection systems.



\bibliography{example_paper}

@article{liu2020primer,
  title={A primer on zeroth-order optimization in signal processing and machine learning: Principals, recent advances, and applications},
  author={Liu, Sijia and Chen, Pin-Yu and Kailkhura, Bhavya and Zhang, Gaoyuan and Hero III, Alfred O and Varshney, Pramod K},
  journal={IEEE Signal Processing Magazine},
  volume={37},
  number={5},
  pages={43--54},
  year={2020},
  publisher={IEEE}
}

@inproceedings{lipton2014optimal,
  title={Optimal thresholding of classifiers to maximize F1 measure},
  author={Lipton, Zachary C and Elkan, Charles and Naryanaswamy, Balakrishnan},
  booktitle={Machine Learning and Knowledge Discovery in Databases: European Conference, ECML PKDD 2014, Nancy, France, September 15-19, 2014. Proceedings, Part II 14},
  pages={225--239},
  year={2014},
  organization={Springer}
}

@inproceedings{feldman2015certifying,
  title={Certifying and removing disparate impact},
  author={Feldman, Michael and Friedler, Sorelle A and Moeller, John and Scheidegger, Carlos and Venkatasubramanian, Suresh},
  booktitle={proceedings of the 21th ACM SIGKDD international conference on knowledge discovery and data mining},
  pages={259--268},
  year={2015}
}

@inproceedings{menon2018cost,
  title={The cost of fairness in binary classification},
  author={Menon, Aditya Krishna and Williamson, Robert C},
  booktitle={Conference on Fairness, accountability and transparency},
  pages={107--118},
  year={2018},
  organization={PMLR}
}

@inproceedings{corbett2017algorithmic,
  title={Algorithmic decision making and the cost of fairness},
  author={Corbett-Davies, Sam and Pierson, Emma and Feller, Avi and Goel, Sharad and Huq, Aziz},
  booktitle={Proceedings of the 23rd acm sigkdd international conference on knowledge discovery and data mining},
  pages={797--806},
  year={2017}
}

@inproceedings{li-etal-2024-mage,
    title = "{MAGE}: Machine-generated Text Detection in the Wild",
    author = "Li, Yafu  and
      Li, Qintong  and
      Cui, Leyang  and
      Bi, Wei  and
      Wang, Zhilin  and
      Wang, Longyue  and
      Yang, Linyi  and
      Shi, Shuming  and
      Zhang, Yue",
    editor = "Ku, Lun-Wei  and
      Martins, Andre  and
      Srikumar, Vivek",
    booktitle = "Proceedings of the 62nd Annual Meeting of the Association for Computational Linguistics (Volume 1: Long Papers)",
    month = aug,
    year = "2024",
    address = "Bangkok, Thailand",
    publisher = "Association for Computational Linguistics",
    doi = "10.18653/v1/2024.acl-long.3",
    pages = "36--53",
}

@inproceedings{jang2022group,
  title={Group-aware threshold adaptation for fair classification},
  author={Jang, Taeuk and Shi, Pengyi and Wang, Xiaoqian},
  booktitle={Proceedings of the AAAI Conference on Artificial Intelligence},
  volume={36},
  number={6},
  pages={6988--6995},
  year={2022}
}

@inproceedings{mitchell2023detectgpt,
  title={Detectgpt: Zero-shot machine-generated text detection using probability curvature},
  author={Mitchell, Eric and Lee, Yoonho and Khazatsky, Alexander and Manning, Christopher D and Finn, Chelsea},
  booktitle={International Conference on Machine Learning},
  pages={24950--24962},
  year={2023},
  organization={PMLR}
}

@article{gehrmann2019gltr,
  title={Gltr: Statistical detection and visualization of generated text},
  author={Gehrmann, Sebastian and Strobelt, Hendrik and Rush, Alexander M},
  journal={arXiv preprint arXiv:1906.04043},
  year={2019}
}

@article{wu2023survey,
  title={A survey on llm-gernerated text detection: Necessity, methods, and future directions},
  author={Wu, Junchao and Yang, Shu and Zhan, Runzhe and Yuan, Yulin and Wong, Derek F and Chao, Lidia S},
  journal={arXiv preprint arXiv:2310.14724},
  year={2023}
}

@article{chen2024combating,
  title={Combating misinformation in the age of llms: Opportunities and challenges},
  author={Chen, Canyu and Shu, Kai},
  journal={AI Magazine},
  volume={45},
  number={3},
  pages={354--368},
  year={2024},
  publisher={Wiley Online Library}
}

@article{guo-etal-2023-hc3,
    title = "How Close is ChatGPT to Human Experts? Comparison Corpus, Evaluation, and Detection",
    author = "Guo, Biyang  and
      Zhang, Xin  and
      Wang, Ziyuan  and
      Jiang, Minqi  and
      Nie, Jinran  and
      Ding, Yuxuan  and
      Yue, Jianwei  and
      Wu, Yupeng",
    journal={arXiv preprint arxiv:2301.07597},
    year = "2023",
}

@misc{desklib-ai-text-detector-2024,
    title = "{DeskLib AI Text Detector v1.01}",
    author = "{DeskLib}",
    howpublished = "\href{https://huggingface.co/desklib/ai-text-detector-v1.01}{https://huggingface.co/desklib/ai-text-detector-v1.01}",
    year = "2024",
    note = "Hugging Face model repository"
}

@misc{openai-gpt4.1-nano-2025,
    title = "{Introducing GPT-4.1 in the API}",
    author = "{OpenAI}",
    howpublished = "\href{https://openai.com/index/gpt-4-1/}{https://openai.com/index/gpt-4-1/}",
    year = "2025",
    note = "Accessed via OpenAI API"
}

@misc{openai-2024-gpt4o-mini,
    title = "{GPT-4o mini: advancing cost-efficient intelligence}",
    author = "{OpenAI}",
    howpublished = "\href{https://openai.com/index/gpt-4o-mini-advancing-cost-efficient-intelligence/}{https://openai.com/index/gpt-4o-mini-advancing-cost-efficient-intelligence/}",
    year = "2024",
    note = "Accessed via OpenAI API"
}

@article{guo2024biscope,
  title={BiScope: AI-generated Text Detection by Checking Memorization of Preceding Tokens},
  author={Guo, Hanxi and Cheng, Siyuan and Jin, Xiaolong and Zhang, Zhuo and Zhang, Kaiyuan and Tao, Guanhong and Shen, Guangyu and Zhang, Xiangyu},
  journal={Advances in Neural Information Processing Systems},
  volume={37},
  pages={104065--104090},
  year={2024}
}

@article{wang2023aleatoric,
  title={Aleatoric and epistemic discrimination: Fundamental limits of fairness interventions},
  author={Wang, Hao and He, Luxi and Gao, Rui and Calmon, Flavio},
  journal={Advances in Neural Information Processing Systems},
  volume={36},
  pages={27040--27062},
  year={2023}
}

@misc{zhou2024e5smalllora,
  author       = {May Zhou},
  title        = {{e5-small-lora-ai-generated-detector}},
  year         = {2024},
  howpublished = {\href{https://huggingface.co/MayZhou/e5-small-lora-ai-generated-detector}{https://huggingface.co/MayZhou/e5-small-lora-ai-generated-detector}},
  note         = {Accessed: 2025-05-09}
}

@inproceedings{dugan-etal-2024-raid,
    title = "{RAID}: A Shared Benchmark for Robust Evaluation of Machine-Generated Text Detectors",
    author = "Dugan, Liam  and
      Hwang, Alyssa  and
      Trhl{\'\i}k, Filip  and
      Zhu, Andrew  and
      Ludan, Josh Magnus  and
      Xu, Hainiu  and
      Ippolito, Daphne  and
      Callison-Burch, Chris",
    booktitle = "Proceedings of the 62nd Annual Meeting of the Association for Computational Linguistics (Volume 1: Long Papers)",
    month = aug,
    year = "2024",
    address = "Bangkok, Thailand",
    publisher = "Association for Computational Linguistics",
    pages = "12463--12492",
}

@article{alghamdi2022beyond,
  title={Beyond adult and compas: Fair multi-class prediction via information projection},
  author={Alghamdi, Wael and Hsu, Hsiang and Jeong, Haewon and Wang, Hao and Michalak, Peter and Asoodeh, Shahab and Calmon, Flavio},
  journal={Advances in Neural Information Processing Systems},
  volume={35},
  pages={38747--38760},
  year={2022}
}

@article{bendekgey2021scalable,
  title={Scalable and stable surrogates for flexible classifiers with fairness constraints},
  author={Bendekgey, Henry C and Sudderth, Erik},
  journal={Advances in Neural Information Processing Systems},
  volume={34},
  pages={30023--30036},
  year={2021}
}

@article{weber2020black,
  title={Black loans matter: Distributionally robust fairness for fighting subgroup discrimination},
  author={Weber, Mark and Yurochkin, Mikhail and Botros, Sherif and Markov, Vanio},
  journal={arXiv preprint arXiv:2012.01193},
  year={2020}
}

@inproceedings{wang2024semeval,
    title = "M4: Multi-generator, Multi-domain, and Multi-lingual Black-Box Machine-Generated Text Detection",
    author = "Wang, Yuxia  and
      Mansurov, Jonibek  and
      Ivanov, Petar  and
      Su, Jinyan  and
      Shelmanov, Artem  and
      Tsvigun, Akim  and
      Whitehouse, Chenxi  and
      Mohammed Afzal, Osama  and
      Mahmoud, Tarek  and
      Sasaki, Toru  and
      Arnold, Thomas  and
      Aji, Alham  and
      Habash, Nizar  and
      Gurevych, Iryna  and
      Nakov, Preslav",
    editor = "Graham, Yvette  and
      Purver, Matthew",
    booktitle = "Proceedings of the 18th Conference of the European Chapter of the Association for Computational Linguistics (Volume 1: Long Papers)",
    month = mar,
    year = "2024",
    address = "St. Julian{'}s, Malta",
    publisher = "Association for Computational Linguistics",
    pages = "1369--1407",
}

@article{hardt2016equality,
  title={Equality of opportunity in supervised learning},
  author={Hardt, Moritz and Price, Eric and Srebro, Nati},
  journal={Advances in neural information processing systems},
  volume={29},
  year={2016}
}

@inproceedings{kim2020fact,
  title={FACT: A diagnostic for group fairness trade-offs},
  author={Kim, Joon Sik and Chen, Jiahao and Talwalkar, Ameet},
  booktitle={International Conference on Machine Learning},
  pages={5264--5274},
  year={2020},
  organization={PMLR}
}

@inproceedings{dwork2012fairness,
  title={Fairness through awareness},
  author={Dwork, Cynthia and Hardt, Moritz and Pitassi, Toniann and Reingold, Omer and Zemel, Richard},
  booktitle={Proceedings of the 3rd innovations in theoretical computer science conference},
  pages={214--226},
  year={2012}
}

@inproceedings{bakker2021beyond,
  title={Beyond reasonable doubt: Improving fairness in budget-constrained decision making using confidence thresholds},
  author={Bakker, Michiel A and Tu, Duy Patrick and Gummadi, Krishna P and Pentland, Alex Sandy and Varshney, Kush R and Weller, Adrian},
  booktitle={Proceedings of the 2021 AAAI/ACM Conference on AI, Ethics, and Society},
  pages={346--356},
  year={2021}
}

@inproceedings{canetti2019soft,
  title={From soft classifiers to hard decisions: How fair can we be?},
  author={Canetti, Ran and Cohen, Aloni and Dikkala, Nishanth and Ramnarayan, Govind and Scheffler, Sarah and Smith, Adam},
  booktitle={Proceedings of the conference on fairness, accountability, and transparency},
  pages={309--318},
  year={2019}
}

@misc{openai2023classifier,
  author = {OpenAI},
  title = {New AI Classifier for Indicating AI-Written Text},
  year = {2023},
   howpublished={\href{https://openai.com/index/new-ai-classifier-for-indicating-ai-written-text}{https://openai.com/index/new-ai-classifier-for-indicating-ai-written-text/} },
  note = {Published: 2023-01-31}
}

@article{Formality_Heylighen2002,
  author    = {Francis Heylighen and
               Jean-Marc Dewaele},
  title     = {Variation in the Contextuality of Language: An Empirical Measure},
  journal   = {Foundations of Science},
  volume    = {7},
  pages     = {293--340},
  year      = {2002},
  doi       = {10.1023/A:1019661126744},
  url       = {https://doi.org/10.1023/A:1019661126744}
}

@article{DistilBERTAD2019,
  title={DistilBERT, a distilled version of BERT: smaller, faster, cheaper and lighter},
  author={Victor Sanh and Lysandre Debut and Julien Chaumond and Thomas Wolf},
  journal={ArXiv},
  year={2019},
  volume={abs/1910.01108}
}

@misc{personality_prediction,
  author = {Minej},
  title = {BERT-based Personality Prediction},
  howpublished = {\href{https://huggingface.co/Minej/bert-base-personality}{https://huggingface.co/Minej/bert-base-personality}},
  note = {Model updated on July 13, 2023},
  year = "2023"
}

@article{bhattacharjee2024fighting,
  title={Fighting fire with fire: can ChatGPT detect AI-generated text?},
  author={Bhattacharjee, Amrita and Liu, Huan},
  journal={ACM SIGKDD Explorations Newsletter},
  volume={25},
  number={2},
  pages={14--21},
  year={2024},
  publisher={ACM New York, NY, USA}
}

@article{solaiman2019release,
  title={Release strategies and the social impacts of language models},
  author={Solaiman, Irene and Brundage, Miles and Clark, Jack and Askell, Amanda and Herbert-Voss, Ariel and Wu, Jeff and Radford, Alec and Krueger, Gretchen and Kim, Jong Wook and Kreps, Sarah and others},
  journal={arXiv preprint arXiv:1908.09203},
  year={2019}
}

@article{bao2023fast,
  title={Fast-detectgpt: Efficient zero-shot detection of machine-generated text via conditional probability curvature},
  author={Bao, Guangsheng and Zhao, Yanbin and Teng, Zhiyang and Yang, Linyi and Zhang, Yue},
  journal={arXiv preprint arXiv:2310.05130},
  year={2023}
}

@article{hu2023radar,
  title={Radar: Robust ai-text detection via adversarial learning},
  author={Hu, Xiaomeng and Chen, Pin-Yu and Ho, Tsung-Yi},
  journal={Advances in Neural Information Processing Systems},
  volume={36},
  pages={15077--15095},
  year={2023}
}

@article{ferrer2022analysis,
  title={Analysis and comparison of classification metrics},
  author={Ferrer, Luciana},
  journal={arXiv preprint arXiv:2209.05355},
  year={2022}
}

@article{perkins2023academic,
  title={Academic Integrity considerations of AI Large Language Models in the post-pandemic era: ChatGPT and beyond},
  author={Perkins, Mike},
  journal={Journal of University Teaching and Learning Practice},
  volume={20},
  number={2},
  year={2023}
}

@article{yao2024survey,
  title={A survey on large language model (llm) security and privacy: The good, the bad, and the ugly},
  author={Yao, Yifan and Duan, Jinhao and Xu, Kaidi and Cai, Yuanfang and Sun, Zhibo and Zhang, Yue},
  journal={High-Confidence Computing},
  pages={100211},
  year={2024},
  publisher={Elsevier}
}

@article{freeman2008comparison,
  title={A comparison of the performance of threshold criteria for binary classification in terms of predicted prevalence and kappa},
  author={Freeman, Elizabeth A and Moisen, Gretchen G},
  journal={Ecological modelling},
  volume={217},
  number={1-2},
  pages={48--58},
  year={2008},
  publisher={Elsevier}
}

@article{manel1999comparing,
  title={Comparing discriminant analysis, neural networks and logistic regression for predicting species distributions: a case study with a Himalayan river bird},
  author={Manel, St{\'e}phanie and Dias, Jean-Marie and Ormerod, Steve J},
  journal={Ecological modelling},
  volume={120},
  number={2-3},
  pages={337--347},
  year={1999},
  publisher={Elsevier}
}

@article{shao2016characterizing,
  title={Characterizing major agricultural land change trends in the Western Corn Belt},
  author={Shao, Yang and Taff, Gregory N and Ren, Jie and Campbell, James B},
  journal={ISPRS Journal of Photogrammetry and Remote Sensing},
  volume={122},
  pages={116--125},
  year={2016},
  publisher={Elsevier}
}

@article{hanberry2013prevalence,
  title={Prevalence, statistical thresholds, and accuracy assessment for species distribution models},
  author={Hanberry, BB and He, HS},
  journal={Web Ecology},
  volume={13},
  number={1},
  pages={13--19},
  year={2013},
  publisher={Copernicus Publications G{\"o}ttingen, Germany}
}

@article{wu2025survey,
  title={A survey on LLM-generated text detection: Necessity, methods, and future directions},
  author={Wu, Junchao and Yang, Shu and Zhan, Runzhe and Yuan, Yulin and Chao, Lidia Sam and Wong, Derek Fai},
  journal={Computational Linguistics},
  pages={1--66},
  year={2025},
  publisher={MIT Press 255 Main Street, 9th Floor, Cambridge, Massachusetts 02142, USA~…}
}

@inproceedings{lu2024mlnet,
  title={MLNet: Mutual Learning Network with Neighborhood Invariance for Universal Domain Adaptation},
  author={Lu, Yanzuo and Shen, Meng and Ma, Andy J and Xie, Xiaohua and Lai, Jian-Huang},
  booktitle={Proceedings of the AAAI Conference on Artificial Intelligence},
  volume={38},
  number={4},
  pages={3900--3908},
  year={2024}
}

@article{esposito2021ghost,
  title={GHOST: adjusting the decision threshold to handle imbalanced data in machine learning},
  author={Esposito, Carmen and Landrum, Gregory A and Schneider, Nadine and Stiefl, Nikolaus and Riniker, Sereina},
  journal={Journal of Chemical Information and Modeling},
  volume={61},
  number={6},
  pages={2623--2640},
  year={2021},
  publisher={ACS Publications}
}

@article{stavnkova2023threshold,
  title={Threshold moving approach with logit models for bankruptcy prediction},
  author={Sta{\v{n}}kov{\'a}, Michaela},
  journal={Computational Economics},
  volume={61},
  number={3},
  pages={1251--1272},
  year={2023},
  publisher={Springer}
}

@article{bradley1997use,
  title={The use of the area under the ROC curve in the evaluation of machine learning algorithms},
  author={Bradley, Andrew P},
  journal={Pattern recognition},
  volume={30},
  number={7},
  pages={1145--1159},
  year={1997},
  publisher={Elsevier}
}

@article{krishna2024paraphrasing,
  title={Paraphrasing evades detectors of ai-generated text, but retrieval is an effective defense},
  author={Krishna, Kalpesh and Song, Yixiao and Karpinska, Marzena and Wieting, John and Iyyer, Mohit},
  journal={Advances in Neural Information Processing Systems},
  volume={36},
  year={2024}
}

@inproceedings{kleinberg2017inherent,
  title={Inherent Trade-Offs in the Fair Determination of Risk Scores},
  author={Kleinberg, Jon and Mullainathan, Sendhil and Raghavan, Manish},
  booktitle={Proceedings of Innovations in Theoretical Computer Science (ITCS)},
  year={2017}
}
\bibliographystyle{icml2026}

\newpage
\appendix
\onecolumn


\section{Performance and fairness metrics} \label{app:performance_fairness_metrics}


\subsection{Contingency Table}
\label{app:contingency_table}

A contingency table (also called a confusion matrix) is a cross-tabulation that summarizes the relationship between true labels and predicted labels in binary classification. It displays four mutually exclusive outcomes:

\begin{itemize}
    \item \textbf{True Positive (TP)}: $Y=1, \hat{Y}=1$. Correct positive prediction.
    \item \textbf{True Negative (TN)}: $Y=0, \hat{Y}=0$. Correct negative prediction.
    \item \textbf{False Positive (FP)}: $Y=0, \hat{Y}=1$. Incorrect positive prediction.
    \item \textbf{False Negative (FN)}: $Y=1, \hat{Y}=0$. Incorrect negative prediction.
\end{itemize}

In the context of AI-text detection, these outcomes are presented in Table~\ref{tab:contingency_matrix}:

\begin{table}[H]
    \centering
    \begin{tabularx}{\linewidth}{|p{0.2\linewidth}|X|X|}
        \hline
        & Classified as AI (\(\hat{Y_1}\)) & Classified as non-AI (\(\hat{Y_0}\)) \\
        \hline
        AI-generated ($Y_1$) & \textbf{TP}: Correct identification & \textbf{FN}: Failed to detect AI and misidentified as human\\
        \hline
        Human-developed ($Y_0$) & \textbf{FP}: Incorrectly marked human work as AI & \textbf{TN}: Correctly identified by not classifying human work as AI \\
        \hline
    \end{tabularx}
    \caption{Contingency matrix for AI-text detection.}
    \label{tab:contingency_matrix}
\end{table}


\subsection{Metrics}
\label{app:metrics}


\paragraph{Accuracy (ACC).}
Accuracy measures the fraction of correct predictions across all outcomes.

\begin{equation}
    \mathrm{ACC} = \frac{\mathrm{TP} + \mathrm{TN}}{\mathrm{TP} + \mathrm{TN} + \mathrm{FP} + \mathrm{FN}}.
\end{equation}


\paragraph{Precision.}
Precision measures the fraction of predicted positives that are correct.

\begin{equation}
    \mathrm{Precision} = \frac{\mathrm{TP}}{\mathrm{TP} + \mathrm{FP}}.
\end{equation}


\paragraph{Recall (TPR).}
Recall measures the fraction of actual positives correctly identified by the model.

\begin{equation}
    \mathrm{Recall} = \mathrm{TPR} = \frac{\mathrm{TP}}{\mathrm{TP} + \mathrm{FN}} = P(\hat{Y}=1 \mid Y=1) 
\end{equation}


\paragraph{F1 Score.}
F1 balances precision and recall into a single metric, particularly useful for imbalanced or multi-objective scenarios.

\begin{equation}
    \mathrm{F1} = 2 \times \frac{\mathrm{Precision} \times \mathrm{Recall}}{\mathrm{Precision} + \mathrm{Recall}}.
\end{equation}


\paragraph{False Positive Rate (FPR).} False Positive Rate measures the proportion of actual negatives incorrectly classified as positive, capturing false alarm rates.

\begin{equation}
    \mathrm{FPR} = \frac{\mathrm{FP}}{\mathrm{FP} + \mathrm{TN}} = P(\hat{Y}=1 \mid Y=0)
\end{equation}


\paragraph{Specificity (True Negative Rate).}

Specificity measures the proportion of actual negatives correctly identified, capturing the model's ability to avoid false alarms.

\begin{equation}
    \mathrm{Specificity} = \mathrm{TNR} = \frac{\mathrm{TN}}{\mathrm{TN} + \mathrm{FP}}.
\end{equation}


\subsection{Impossibility Theorem}
\label{sec:impossibility}

The impossibility theorem in algorithmic fairness establishes a fundamental constraint: no single classifier can simultaneously satisfy all common fairness metrics (e.g., demographic parity, equalized odds, predictive parity) except in trivial cases. This incompatibility arises because different fairness definitions impose mathematically conflicting constraints when base rates (class prevalences) differ across protected groups.


\section{Applied Thresholds}
\label{app:applied_thresholds}

We evaluate three thresholding strategies: static ($\theta = 0.5$), AUROC-based ($\theta = 0.2866$), and adaptive per-subgroup thresholds learned by \textit{\name}. Table~\ref{tab:applied_thresholds_60} summarizes the results across 60 subgroups defined by four attributes: text length, personality trait, formality, and sentiment polarity.

\begin{figure*}[h!]
    \centering
    \includegraphics[width=0.92\textwidth]{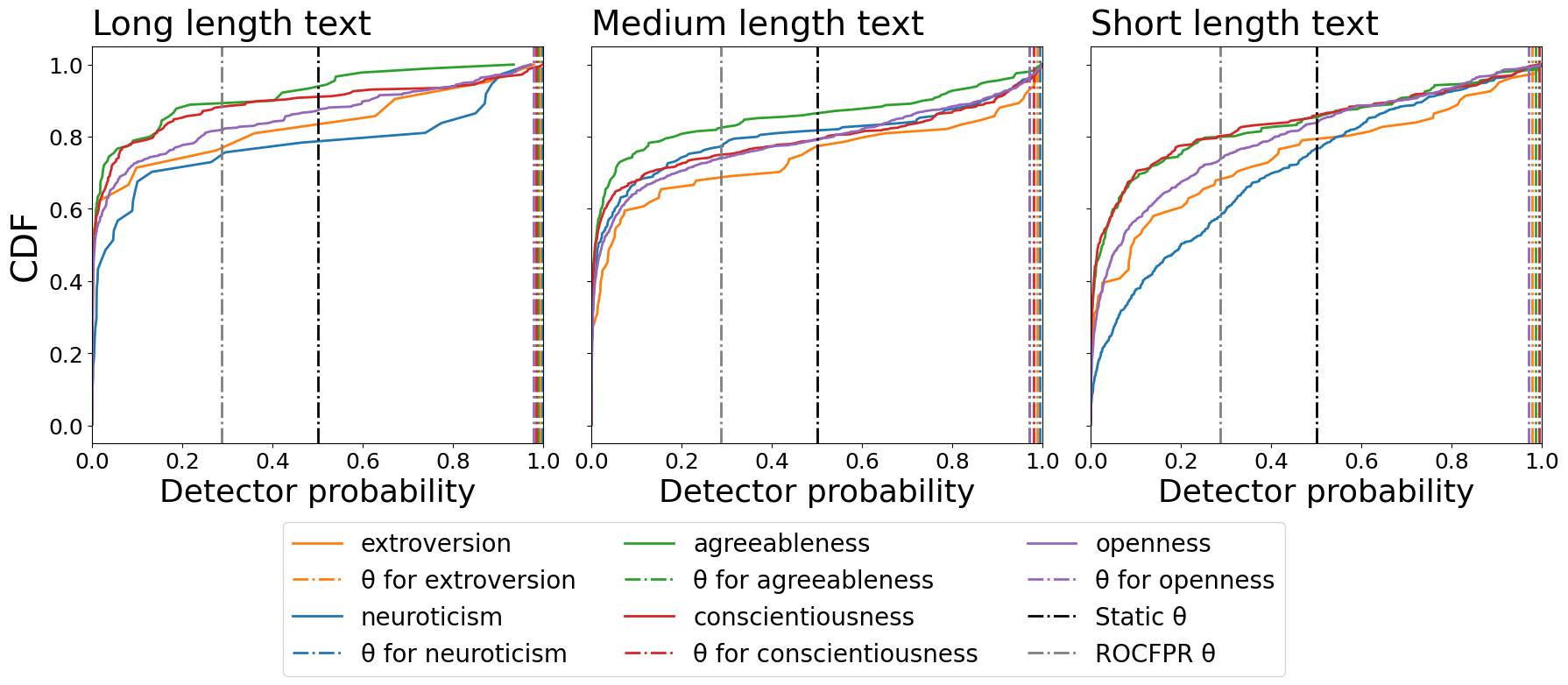}
    \caption{Cumulative density functions (CDFs) of probability distributions by subgroup (text length and personality traits from Table~\ref{tab:featuretable}). Static ($\theta=0.5$, black) and AUROC-based (gray) thresholds do not account for subgroup-specific variations, while \textit{\name} learns adaptive thresholds for each of the 15 subgroups. Trained with RoBERTa-large detector.}
    \label{fig:application_probability}
\end{figure*}

Figure~\ref{fig:application_probability} visualizes these strategies for a two-attribute subset (text length and personality) to illustrate the trade-off between fairness and performance. The full results in Table~\ref{tab:applied_thresholds_15} demonstrate that the adaptive approach generalizes to multi-dimensional subgroup definitions, yielding heterogeneous thresholds that account for fine-grained distributional differences.

\begin{longtable}{|l|l|c|}
    \hline
    \multicolumn{3}{|c|}{\textbf{Adaptive Thresholds ($n$ = 15 subgroups)}} \\
    \hline
    \multicolumn{3}{|c|}{\textbf{Static: 0.5 $|$ ROCFPR: 0.2866}} \\
    \hline
    \textbf{Text Length} & \textbf{Writing style} & \textbf{Threshold (\emph{\name})} \\
    \hline
    \endhead
    
    \hline
    \multicolumn{3}{r}{\textit{Continued on next page}} \\
    \endfoot
    
    \endlastfoot
    
    long & agreeableness & 0.9881 \\
    long & conscientiousness & 0.9832 \\
    long & extroversion & 0.9915 \\
    long & neuroticism & 0.9983 \\
    long & openness & 0.9783 \\
    medium & agreeableness & 0.9712 \\
    medium & conscientiousness & 0.9808 \\
    medium & extroversion & 0.9875 \\
    medium & neuroticism & 0.9943 \\
    medium & openness & 0.9698 \\
    short & agreeableness & 0.9865 \\
    short & conscientiousness & 0.9942 \\
    short & extroversion & 0.9781 \\
    short & neuroticism & 0.9942 \\
    short & openness & 0.9707 \\
    \hline
    \caption{Adaptive thresholds learned by \emph{\name} across 15 subgroups defined by text length and writing style.}
    \label{tab:applied_thresholds_15}
\end{longtable}

\begin{longtable}{|l|l|l|l|c|}
    \hline
    \multicolumn{5}{|c|}{\textbf{Adaptive Thresholds ($n$ = 60 subgroups)}} \\
    \hline
    \multicolumn{5}{|c|}{\textbf{Static: 0.5 $|$ ROCFPR: 0.2866}} \\
    \hline
    \textbf{Text Length} & \textbf{Writing style} & \textbf{Formality} & \textbf{Sentiment} & \textbf{Threshold (\emph{\name})} \\
    \hline
    \endhead
    
    \hline
    \multicolumn{5}{r}{\textit{Continued on next page}} \\
    \endfoot
    
    \endlastfoot
    
    short & neuroticism & formal & negative & 0.9914 \\
    short & neuroticism & formal & positive & 0.9893 \\
    short & neuroticism & informal & negative & 0.9835 \\
    short & neuroticism & informal & positive & 0.9851 \\
    short & extroversion & formal & negative & 0.9797 \\
    short & extroversion & formal & positive & 0.9772 \\
    short & extroversion & informal & negative & 0.9645 \\
    short & extroversion & informal & positive & 0.9668 \\
    short & agreeableness & formal & negative & 0.9360 \\
    short & agreeableness & formal & positive & 0.9385 \\
    short & agreeableness & informal & negative & 0.9768 \\
    short & agreeableness & informal & positive & 0.9741 \\
    short & conscientiousness & formal & negative & 0.9944 \\
    short & conscientiousness & formal & positive & 0.9926 \\
    short & conscientiousness & informal & negative & 0.9810 \\
    short & conscientiousness & informal & positive & 0.9823 \\
    short & openness & formal & negative & 0.9724 \\
    short & openness & formal & positive & 0.9747 \\
    short & openness & informal & negative & 0.9676 \\
    short & openness & informal & positive & 0.9692 \\
    \hline
    medium & neuroticism & formal & negative & 0.9924 \\
    medium & neuroticism & formal & positive & 0.9901 \\
    medium & neuroticism & informal & negative & 0.9881 \\
    medium & neuroticism & informal & positive & 0.9865 \\
    medium & extroversion & formal & negative & 0.9889 \\
    medium & extroversion & formal & positive & 0.9902 \\
    medium & extroversion & informal & negative & 0.9736 \\
    medium & extroversion & informal & positive & 0.9751 \\
    medium & agreeableness & formal & negative & 0.9714 \\
    medium & agreeableness & formal & positive & 0.9738 \\
    medium & agreeableness & informal & negative & 0.9665 \\
    medium & agreeableness & informal & positive & 0.9682 \\
    medium & conscientiousness & formal & negative & 0.9825 \\
    medium & conscientiousness & formal & positive & 0.9846 \\
    medium & conscientiousness & informal & negative & 0.9776 \\
    medium & conscientiousness & informal & positive & 0.9791 \\
    medium & openness & formal & negative & 0.9758 \\
    medium & openness & formal & positive & 0.9775 \\
    medium & openness & informal & negative & 0.9698 \\
    medium & openness & informal & positive & 0.9712 \\
    \hline
    long & neuroticism & formal & negative & 0.9954 \\
    long & neuroticism & formal & positive & 0.9932 \\
    long & neuroticism & informal & negative & 0.9937 \\
    long & neuroticism & informal & positive & 0.9951 \\
    long & extroversion & formal & negative & 0.9957 \\
    long & extroversion & formal & positive & 0.9972 \\
    long & extroversion & informal & negative & 0.9897 \\
    long & extroversion & informal & positive & 0.9915 \\
    long & agreeableness & formal & negative & 0.9835 \\
    long & agreeableness & formal & positive & 0.9851 \\
    long & agreeableness & informal & negative & 0.9776 \\
    long & agreeableness & informal & positive & 0.9798 \\
    long & conscientiousness & formal & negative & 0.9887 \\
    long & conscientiousness & formal & positive & 0.9902 \\
    long & conscientiousness & informal & negative & 0.9714 \\
    long & conscientiousness & informal & positive & 0.9735 \\
    long & openness & formal & negative & 0.9754 \\
    long & openness & formal & positive & 0.9771 \\
    long & openness & informal & negative & 0.9680 \\
    long & openness & informal & positive & 0.9702 \\
    \hline
    
    \caption{Adaptive thresholds learned by \emph{\name} across 60 subgroups defined by text length, writing style, formality, and sentiment.}
    \label{tab:applied_thresholds_60}
\end{longtable}


\section{Detailed Results}
\label{app:detailed_results}

\subsection{Overall discrepancy and performance}
Table \ref{overall_table} summarizes the overall discrepancy and performance by dataset and thresholding method.

\begin{table}[ht]
\centering
\small
\begin{tabular}{llccc}
\toprule
Source    & Strategy & Discrepancy (mean\,$\pm$\,std) & Accuracy (mean\,$\pm$\,std) & F1 Score (mean\,$\pm$\,std) \\
\midrule[0.8pt]
MAGE      & FairOPT  & $\mathbf{0.4495\pm0.1656}$ & $0.6339\pm0.1512$ & $0.3877\pm0.3494$ \\
          & ROCFPR   & $0.5551\pm0.1664$ & $0.6238\pm0.2113$ & $0.5326\pm0.3170$ \\
          & Static   & $0.5433\pm0.1781$ & $0.6388\pm0.2015$ & $0.5126\pm0.3352$ \\
\midrule
RAID      & FairOPT  & $\mathbf{0.4836\pm0.1781}$ & $0.6280\pm0.1573$ & $0.3885\pm0.3535$ \\
          & ROCFPR   & $0.5644\pm0.1263$ & $0.6253\pm0.2082$ & $0.5363\pm0.3185$ \\
          & Static   & $0.5562\pm0.1478$ & $0.6364\pm0.1994$ & $0.5136\pm0.3358$ \\
\midrule
SemEval   & FairOPT  & $\mathbf{0.3763\pm0.2166}$ & $0.6404\pm0.1497$ & $0.3896\pm0.3484$ \\
          & ROCFPR   & $0.5598\pm0.1758$ & $0.6258\pm0.2010$ & $0.5303\pm0.3114$ \\
          & Static   & $0.5548\pm0.1969$ & $0.6376\pm0.1948$ & $0.5081\pm0.3288$ \\
\bottomrule
\end{tabular}
\caption{Overall discrepancy and performance by data source and thresholding strategy (mean\,$\pm$\,std)}
\label{overall_table}
\end{table}

\subsection{Comparison by performance and fairness metrics by datasource, detector, and thresholding strategy}

The results are presented by each detector based on fairness and performance metrics from Table \ref{tab:overall_by_detector_and_dataset} (mean $\pm$ std).

\begin{longtable}{lllccc}

\midrule
\multicolumn{6}{r}{\textit{Continued on next page}} \\
\endfoot

\bottomrule
\endlastfoot

\midrule
\multicolumn{6}{l}{\textbf{Dataset}} \\
& Detector    & Thresholding  & Discrepancy & ACC & F1 score \\

\midrule
\midrule
\multicolumn{6}{l}{\textbf{MAGE}} \\
& GPT4.1-nano (2025)    & FairOPT  & $0.4106\pm0.2350$ & $0.4794\pm0.0374$ & $0.0114\pm0.0121$ \\
&                       & ROCFPR   & $0.4636\pm0.2194$ & $0.4274\pm0.0380$ & $0.0694\pm0.0159$ \\
&                       & Static   & $0.4647\pm0.2143$ & $0.4393\pm0.0401$ & $0.0507\pm0.0154$ \\

\midrule[0.3pt]
& GPT4o-mini (2024)     & FairOPT  & $0.3820\pm0.2421$ & $0.4866\pm0.0345$ & $0.0046\pm0.0078$ \\
&                       & ROCFPR   & $0.6874\pm0.1882$ & $0.4453\pm0.0288$ & $0.2348\pm0.0399$ \\
&                       & Static   & $0.6517\pm0.1661$ & $0.4423\pm0.0253$ & $0.1947\pm0.0304$ \\

\midrule[0.3pt]
& SimpleAI (2023)       & FairOPT  & $0.4536\pm0.0871$ & $0.7501\pm0.0274$ & $0.7006\pm0.0363$ \\
&                       & ROCFPR   & $0.5682\pm0.0945$ & $0.7573\pm0.0130$ & $0.7520\pm0.0215$ \\
&                       & Static   & $0.5527\pm0.0968$ & $0.7525\pm0.0154$ & $0.7360\pm0.0298$ \\

\midrule[0.3pt]
& DeskLib (2024)        & FairOPT  & $0.4896\pm0.0914$ & $0.8900\pm0.0170$ & $0.8809\pm0.0196$ \\
&                       & ROCFPR   & $0.4854\pm0.2388$ & $0.8787\pm0.0212$ & $0.8916\pm0.0224$ \\
&                       & Static   & $0.4278\pm0.2561$ & $0.9074\pm0.0206$ & $0.9140\pm0.0219$ \\

\midrule[0.3pt]
& E5LoRA (2024)         & FairOPT  & $0.3571\pm0.2440$ & $0.4895\pm0.0336$ & $0.0000\pm0.0000$ \\
&                       & ROCFPR   & $0.5779\pm0.1866$ & $0.1996\pm0.0302$ & $0.0471\pm0.0179$ \\
&                       & Static   & $0.6026\pm0.2439$ & $0.2797\pm0.0237$ & $0.0065\pm0.0090$ \\

\midrule[0.3pt]
& FastDetectGPT (2023)  & FairOPT  & $0.4929\pm0.0189$ & $0.8099\pm0.0274$ & $0.7774\pm0.0243$ \\
&                       & ROCFPR   & $0.4881\pm0.0315$ & $0.7776\pm0.0302$ & $0.8114\pm0.0288$ \\
&                       & Static   & $0.4911\pm0.1295$ & $0.8398\pm0.0154$ & $0.8520\pm0.0151$ \\

\midrule[0.3pt]
& RoBERTa-base (2019)   & FairOPT  & $0.4893\pm0.0537$ & $0.6742\pm0.0436$ & $0.5381\pm0.0655$ \\
&                       & ROCFPR   & $0.5923\pm0.0884$ & $0.7322\pm0.0311$ & $0.6963\pm0.0320$ \\
&                       & Static   & $0.6158\pm0.1606$ & $0.7256\pm0.0364$ & $0.6722\pm0.0423$ \\

\midrule[0.3pt]
& RoBERTa-large (2019)  & FairOPT  & $0.3940\pm0.1992$ & $0.5063\pm0.0364$ & $0.0656\pm0.0372$ \\
&                       & ROCFPR   & $0.5835\pm0.0610$ & $0.6264\pm0.0289$ & $0.5067\pm0.0405$ \\
&                       & Static   & $0.5414\pm0.1038$ & $0.6103\pm0.0361$ & $0.4315\pm0.0522$ \\

\midrule[0.3pt]
& RADAR (2023)          & FairOPT  & $0.5760\pm0.0755$ & $0.6192\pm0.0216$ & $0.5106\pm0.0388$ \\
&                       & ROCFPR   & $0.5495\pm0.2113$ & $0.7693\pm0.0186$ & $0.7837\pm0.0202$ \\
&                       & Static   & $0.5418\pm0.1280$ & $0.7519\pm0.0224$ & $0.7562\pm0.0242$ \\

\midrule
\midrule
\multicolumn{6}{l}{\textbf{RAID}} \\
& GPT4.1-nano (2025)    & FairOPT  & $0.5208\pm0.2082$ & $0.4738\pm0.0087$ & $0.0128\pm0.0115$ \\
&                       & ROCFPR   & $0.5820\pm0.1623$ & $0.4243\pm0.0189$ & $0.0795\pm0.0295$ \\
&                       & Static   & $0.5781\pm0.1558$ & $0.4306\pm0.0183$ & $0.0465\pm0.0202$ \\

\midrule[0.3pt]
& GPT4o-mini (2024)     & FairOPT  & $0.4781\pm0.2428$ & $0.4710\pm0.0180$ & $0.0000\pm0.0000$ \\
&                       & ROCFPR   & $0.5762\pm0.1567$ & $0.4557\pm0.0266$ & $0.2292\pm0.0542$ \\
&                       & Static   & $0.5816\pm0.1733$ & $0.4529\pm0.0180$ & $0.1875\pm0.0321$ \\

\midrule[0.3pt]
& SimpleAI (2023)       & FairOPT  & $0.5868\pm0.2080$ & $0.7320\pm0.0408$ & $0.6835\pm0.0548$ \\
&                       & ROCFPR   & $0.5655\pm0.1818$ & $0.7452\pm0.0386$ & $0.7502\pm0.0388$ \\
&                       & Static   & $0.6245\pm0.2095$ & $0.7403\pm0.0516$ & $0.7357\pm0.0541$ \\

\midrule[0.3pt]
& DeskLib (2024)        & FairOPT  & $0.3552\pm0.1001$ & $0.9023\pm0.0289$ & $0.8973\pm0.0301$ \\
&                       & ROCFPR   & $0.4578\pm0.0522$ & $0.8667\pm0.0223$ & $0.8843\pm0.0194$ \\
&                       & Static   & $0.4292\pm0.0646$ & $0.8988\pm0.0212$ & $0.9076\pm0.0189$ \\

\midrule[0.3pt]
& E5LoRA (2024)         & FairOPT  & $0.4167\pm0.2041$ & $0.4808\pm0.0153$ & $0.0000\pm0.0000$ \\
&                       & ROCFPR   & $0.6437\pm0.1177$ & $0.2059\pm0.0384$ & $0.0406\pm0.0272$ \\
&                       & Static   & $0.6535\pm0.1981$ & $0.2805\pm0.0323$ & $0.0078\pm0.0097$ \\

\midrule[0.3pt]
& FastDetectGPT (2023)  & FairOPT  & $0.4405\pm0.0560$ & $0.8241\pm0.0203$ & $0.7996\pm0.0230$ \\
&                       & ROCFPR   & $0.5103\pm0.0873$ & $0.7850\pm0.0345$ & $0.8191\pm0.0312$ \\
&                       & Static   & $0.4609\pm0.1409$ & $0.8472\pm0.0247$ & $0.8601\pm0.0228$ \\

\midrule[0.3pt]
& RoBERTa-base (2019)   & FairOPT  & $0.5361\pm0.0452$ & $0.6706\pm0.0343$ & $0.5483\pm0.0681$ \\
&                       & ROCFPR   & $0.6246\pm0.1244$ & $0.7348\pm0.0355$ & $0.7030\pm0.0478$ \\
&                       & Static   & $0.5978\pm0.0984$ & $0.7180\pm0.0342$ & $0.6682\pm0.0499$ \\

\midrule[0.3pt]
& RoBERTa-large (2019)  & FairOPT  & $0.4282\pm0.1927$ & $0.4990\pm0.0183$ & $0.0698\pm0.0285$ \\
&                       & ROCFPR   & $0.6079\pm0.0816$ & $0.6413\pm0.0218$ & $0.5341\pm0.0394$ \\
&                       & Static   & $0.5570\pm0.0658$ & $0.6176\pm0.0175$ & $0.4579\pm0.0319$ \\

\midrule[0.3pt]
& RADAR (2023)          & FairOPT  & $0.5900\pm0.1850$ & $0.5980\pm0.0179$ & $0.4853\pm0.0412$ \\
&                       & ROCFPR   & $0.5113\pm0.0602$ & $0.7683\pm0.0263$ & $0.7867\pm0.0233$ \\
&                       & Static   & $0.5232\pm0.0644$ & $0.7418\pm0.0265$ & $0.7510\pm0.0257$ \\

\midrule
\midrule
\multicolumn{6}{l}{\textbf{SemEval}} \\
& GPT4.1-nano (2025)    & FairOPT  & $0.2663\pm0.2792$ & $0.4889\pm0.0498$ & $0.0110\pm0.0116$ \\
&                       & ROCFPR   & $0.5139\pm0.2100$ & $0.4459\pm0.0354$ & $0.0912\pm0.0320$ \\
&                       & Static   & $0.4889\pm0.2156$ & $0.4495\pm0.0408$ & $0.0503\pm0.0210$ \\

\midrule[0.3pt]
& GPT4o-mini (2024)     & FairOPT  & $0.2763\pm0.2923$ & $0.4943\pm0.0501$ & $0.0029\pm0.0076$ \\
&                       & ROCFPR   & $0.5777\pm0.2485$ & $0.4638\pm0.0350$ & $0.2381\pm0.0484$ \\
&                       & Static   & $0.5617\pm0.2843$ & $0.4620\pm0.0418$ & $0.2008\pm0.0364$ \\

\midrule[0.3pt]
& SimpleAI (2023)       & FairOPT  & $0.5655\pm0.2123$ & $0.7382\pm0.0369$ & $0.6842\pm0.0260$ \\
&                       & ROCFPR   & $0.6726\pm0.1674$ & $0.7430\pm0.0313$ & $0.7375\pm0.0264$ \\
&                       & Static   & $0.6519\pm0.1904$ & $0.7436\pm0.0327$ & $0.7283\pm0.0231$ \\

\midrule[0.3pt]
& DeskLib (2024)        & FairOPT  & $0.3711\pm0.0755$ & $0.8942\pm0.0349$ & $0.8844\pm0.0316$ \\
&                       & ROCFPR   & $0.4487\pm0.0727$ & $0.8643\pm0.0169$ & $0.8776\pm0.0241$ \\
&                       & Static   & $0.3691\pm0.0970$ & $0.8954\pm0.0139$ & $0.9012\pm0.0184$ \\

\midrule[0.3pt]
& E5LoRA (2024)         & FairOPT  & $0.2143\pm0.2673$ & $0.5015\pm0.0523$ & $0.0000\pm0.0000$ \\
&                       & ROCFPR   & $0.6887\pm0.2807$ & $0.2098\pm0.0405$ & $0.0293\pm0.0124$ \\
&                       & Static   & $0.6428\pm0.2790$ & $0.2750\pm0.0402$ & $0.0065\pm0.0061$ \\

\midrule[0.3pt]
& FastDetectGPT (2023)  & FairOPT  & $0.4818\pm0.0312$ & $0.8111\pm0.0254$ & $0.7701\pm0.0298$ \\
&                       & ROCFPR   & $0.4689\pm0.0646$ & $0.7490\pm0.0284$ & $0.7851\pm0.0345$ \\
&                       & Static   & $0.5232\pm0.1360$ & $0.8159\pm0.0253$ & $0.8268\pm0.0276$ \\

\midrule[0.3pt]
& RoBERTa-base (2019)   & FairOPT  & $0.4922\pm0.0426$ & $0.6940\pm0.0371$ & $0.5676\pm0.0472$ \\
&                       & ROCFPR   & $0.5748\pm0.1175$ & $0.7519\pm0.0241$ & $0.7127\pm0.0362$ \\
&                       & Static   & $0.5543\pm0.0866$ & $0.7412\pm0.0263$ & $0.6847\pm0.0435$ \\

\midrule[0.3pt]
& RoBERTa-large (2019)  & FairOPT  & $0.2956\pm0.2093$ & $0.5188\pm0.0513$ & $0.0718\pm0.0086$ \\
&                       & ROCFPR   & $0.5977\pm0.1239$ & $0.6479\pm0.0218$ & $0.5280\pm0.0376$ \\
&                       & Static   & $0.6073\pm0.1346$ & $0.6186\pm0.0293$ & $0.4333\pm0.0302$ \\

\midrule[0.3pt]
& RADAR (2023)          & FairOPT  & $0.4241\pm0.1244$ & $0.6222\pm0.0358$ & $0.5140\pm0.0407$ \\
&                       & ROCFPR   & $0.4956\pm0.0187$ & $0.7561\pm0.0310$ & $0.7735\pm0.0357$ \\
&                       & Static   & $0.5939\pm0.1847$ & $0.7370\pm0.0300$ & $0.7407\pm0.0361$ \\

\midrule[0.8pt]
\caption{Discrepancy and performance (mean\,$\pm$\,std) by data source, detector, and thresholding method}
\label{tab:overall_by_detector_and_dataset}
\end{longtable}


\section{Operationalizing \textit{\name}}

\subsection{Algorithmic description}\label{appendix:algorithm}

Algorithm~\ref{algorithm_fairopt2} presents the FairOPT optimization procedure.


\newlength{\textfloatsepsave}
\setlength{\textfloatsepsave}{\textfloatsep}
\setlength{\textfloatsep}{0pt}
\begin{algorithm}[ht!]
    \begin{algorithmic}[1]
    \footnotesize
    
    \STATE \textbf{Input:} Labeled dataset $D = \{(x_n, y_n)\}$, predicted probabilities $\{p_n\}$, subgroup labels $\{g_n\}$ with possible groups $G_1,\dots,G_j$, features \(\{S_1,\dots,S_n\}\), initial thresholds $\theta_{init}$, fairness metrics $\{M_1, \ldots, M_k\}$ with acceptable disparity $\delta_{\mathrm{fair}}$, minimum performance thresholds: $\alpha$ (ACC) and $\beta$ (F1), penalty weight $\kappa$, learning rate $\eta$, tolerance $\epsilon_{\mathrm{tol}}$, maximum iterations, finite-difference step $\delta$, minimum probability for threshold $a$, maximum probability for threshold $b$

    \STATE \textbf{Output:} Subgroup thresholds $\{\theta^*(G_1),\dots,\theta^*(G_j)\}$
    \STATE \textbf{Initialization:}
    \STATE split $D$ to ${(G_1),\dots,(G_j)}$  by $S$
    \WHILE{\text{iteration} $<$ \text{maximum iterations}}
        \FOR{each group $G_j$}
            \STATE $\mathcal{I}_i \leftarrow \{\,n \mid g_n = G_j\}$ \quad all samples in group $G_j$
            \STATE $\hat{y}_n \leftarrow [\,p_n \ge \theta(G_j)\,], \;\forall n \in \mathcal{I}_i$
            \STATE Store contingency table for $G_i$ using $\hat{y}_n$ and ${y}_n$
            \STATE $\Omega \;\leftarrow\; \min_i\{\mathrm{ACC}(G_j)\} \;\ge\; \alpha \;\wedge\;\min_i\{\mathrm{F1}(G_j)\} \;\ge\; \beta$ boolean output
            \FOR{each fairness metric $M_k$}
                \STATE Compute $\{M_k(G_1),\dots,M_k(G_j)\}$
                \STATE $\Delta_{k} \leftarrow \max_i(M_k(G_j)) - \min_i(M_k(G_j))$
                \STATE $\Psi \;\leftarrow\; \Bigl[\,\max_k \Delta_k \;\le\; \delta_{\mathrm{fair}}\Bigr]$ boolean output
            \ENDFOR
            \FOR{each group $G_j$}
                \STATE $L_i(\theta) = -\mathrm{Acc}_j(\theta) + \kappa\,[\,\beta - \mathrm{F1}_j(\theta)\,]_{+}$
                \STATE
                \[
                    \nabla L_i(\theta(G_j))
                    \;\approx\;
                    \frac{L_i(\theta(G_j) + \delta) \;-\; L_i(\theta(G_j) - \delta)}{2\,\delta}.
                \]
                \STATE
                \[
                    \theta(G_j) \;\leftarrow\; \theta(G_j) \;-\; \eta \;\nabla L_i(\theta(G_j)).
                \]
                \STATE Clip $\theta(G_j)$ to stay within $[a,b]$
            \ENDFOR        
        \ENDFOR
        \STATE $\Delta_\theta \;\leftarrow\; \max_{G_j} \bigl|\theta(G_j) - \text{previousThresholds}[G_j]\bigr|$
        \IF{($\Delta_\theta < \epsilon_{\mathrm{tol}} \;\wedge\; \Omega \;\wedge\; \Psi$)}
            \STATE \textbf{break}
        \ENDIF
    
        \STATE $\text{previousThresholds} \leftarrow \theta(\cdot)$
        \STATE $\text{iteration} \leftarrow \text{iteration}+1$
    \ENDWHILE
    \STATE \textbf{return} $\{\theta(G_1),\dots,\theta(G_j)\}$
    
    \caption{\name: Gradient-based adaptive threshold optimization with fairness for subgroups}
    \label{algorithm_fairopt2}
    \end{algorithmic}
\end{algorithm}
\setlength{\textfloatsep}{\textfloatsepsave}


\subsection{Early stopping method for the tradeoff curve}\label{app:early_stopping}

The \textit{\name} algorithm was implemented using functions from the \texttt{scipy.optimize} library, aiming to determine probability thresholds that satisfy specific fairness constraints \texttt{(acceptable\_disparity =  [1.00, 0.30, 0.25, 0.21, 0.20, 0.19, 0.15, 0.15])} while ensuring minimum performance thresholds for both performance criteria (ACC and F1). The optimization in the \texttt{minimize} function was performed using the \texttt{L-BFGS-B} method. The figure is based on the RoBERTa-base model, and it uses different hyperparameters to get eight results with different relaxation methods. Early stopping is facilitated by the default values of \texttt{ftol} and \texttt{gtol} in \texttt{L-BFGS-B}. Additionally, the \texttt{learning rate} was set to \(10^{-3}\) to enable controlled and stable updates during optimization, and a \texttt{penalty} parameter was set to \(10\) to strengthen the L2 penalty in the loss function. Furthermore, minimum performance thresholds for accuracy and F1-score were enforced based on the performance observed for each group when applying a static threshold of 0.5 and a ROC-based threshold. As stricter fairness constraints are imposed, these minimum thresholds are gradually relaxed. 



\subsection{Hyperparameter settings for the FairOPT}\label{hyperparameters}

The application of the \textit{\name}\ is configured with a set of hyperparameters designed to ensure a balance between fairness and performance during the threshold optimization process. The \texttt{learning rate is set to $10^{-3}$} to enable controlled and stable updates during optimization. \texttt{The maximum number of iterations is specified as $10^5$} to provide sufficient time for the optimization algorithm to converge. An acceptable disparity of 0.2 is defined to regulate subgroup fairness, ensuring that performance differences across subgroups remain within an acceptable range through the relaxed fairness. Minimum \texttt{thresholds for accuracy (0.25) and F1-score (0.25)} are enforced to maintain baseline model performance and stability, particularly in cases of imbalanced datasets. A \texttt{tolerance level of $1/2*10^{-4}$} is introduced to impose convergence criteria, ensuring the optimization process halts only when the changes in the objective function are negligible. Furthermore, a \texttt{penalty parameter of 20} is applied to enforce performance constraints by penalizing significant performance disparities across subgroups. Also, this experiment used early stopping method for effective processing. The early stopping check through the change of the ACC, F1, and maximum DP discrepancy of each group, and terminates if all of them shows minimal change based on predefined patience.


\section{Computational Efficiency}

FairOPT runs in $\mathcal{O}(I \cdot g \cdot n)$ time, where $I$ is the number of iterations, $g$ is the number of subgroups, and $n$ is the dataset size. 
We ran all experiments on one laptop equipped with a 13th-Gen Intel Core i7-13620H (10 cores, 16 threads, 2.40 GHz base, 3.1 GHz turbo) and 16 GB DDR5-4800 MT/s RAM; during optimization on the full training set (19{,}117 samples), CPU utilization averaged 25--30\%, memory usage peaked at approximately 14 GB, and training completed in under six hours.


\end{document}